\documentclass[sigconf]{acmart}

\usepackage{booktabs}
\usepackage{makecell}
\usepackage{amsmath}
\usepackage{amsthm}
\usepackage{pifont}
\usepackage{bbding}
\usepackage{xcolor}
\definecolor{appendixself}{RGB}{35,77,148}
\definecolor{appendixother}{RGB}{148,74,132}
\definecolor{appendixsignal}{RGB}{160,88,24}

\acmYear{2026}
\acmConference[DAI '26]
  {The Eighth International Conference on Distributed Artificial Intelligence}
  {November 29--December 2, 2026}
  {Hong Kong, China}
\acmBooktitle{The Eighth International Conference on Distributed Artificial
  Intelligence (DAI '26), November 29--December 2, 2026, Hong Kong, China}

\setcopyright{none}
\copyrightyear{}
\acmDOI{}
\acmISBN{}

\begin{document}

% Replace all example content with your accepted paper's content.
\title{TradeWar: A Benchmark for Execution-Grounded Trade Negotiation with Adaptive LLM Agents}

% Copy all author blocks from your accepted manuscript here.
% Do not add new authors; the order of existing authors may be changed.
\author{Xiangyu Kong}
\authornote{Both authors contributed equally to this research.\\\Envelope~Corresponding author (E-mail: xykong@bistu.edu.cn).}
\correspondingauthor
\affiliation{%
  \institution{College of Computer Science, Beijing Information Science and Technology University}
  \city{Beijing}
  \country{China}}
\email{xykong@bistu.edu.cn}
% Repeat the author/affiliation/email block for the other existing authors.
% For a long author list, supply a short running author list if needed:
% \renewcommand{\shortauthors}{First Author et al.}
\author{Ruirui Yang}
\authornotemark[1]
\affiliation{%
  \institution{College of Computer Science, Beijing Information Science and Technology University}
  \city{Beijing}
  \country{China}}
\email{yangruirui@bistu.edu.cn}

\author{Xia Hou}
\affiliation{%
  \institution{College of Computer Science, Beijing Information Science and Technology University}
  \city{Beijing}
  \country{China}}
\email{houxia@bistu.edu.cn}

\begin{abstract}
Multi-turn negotiation remains a challenging problem for large language model (LLM) agents. While LLMs can conduct natural-language dialogue, existing benchmarks predominantly assess only agreement rates, overlooking whether negotiated commitments actually create value once executed. To address this gap, we introduce TradeWar, a multi-country simulation benchmark that translates negotiation dialogues into concrete tariff and trade-share policies, enforces capacity and network constraints during settlement, and evaluates outcomes via an Economic Health Index (EHI). To solve this execution-grounded task, we propose TACTIC (Trade-aware Adaptive Consolidation and Theory-of-mind for International Commerce), an agent that iteratively refines its negotiation strategy across episodes---without updating base LLM weights---by leveraging prior experience, opponent behavior, and past settlement outcomes. Despite using zero-shot chain-of-thought (CoT) prompting, PromptLLM attains a 75\% agreement rate but only 57.1 final EHI ($\times 100$), whereas TACTIC reaches 68.8. Across 20 sequential episodes and three seeds, TACTIC's final EHI improves from 58.2 to 73.2, compared with 65.2 for an ablation without cross-episode updates. Under technology-controls and energy-crisis shocks, TACTIC also exhibits smaller performance degradation than PromptLLM and MemoryLLM. These results indicate that CoT deliberation and interaction memory alone do not fully address delayed execution feedback, whereas settlement-grounded adaptation improves long-horizon negotiation in TradeWar\footnote{Code and benchmark: \textcolor{blue}{\textit{\url{https://github.com/kxyzc1992/TradeWar-Benchmark}}}.}.
\end{abstract}

% Paste your paper's own CCSXML and \ccsdesc commands here, if applicable.


% Replace the example keywords with your paper's actual keywords.

\maketitle

\section{Introduction}

Large language model (LLM) agents can negotiate through natural-language dialogue \cite{fu2023improvinglanguage,xia2024measuring,liu2026opponent}, but fluent agreement alone does not establish that a commitment is feasible, binding, or valuable after downstream constraints are applied. This issue becomes more pressing as LLM agents move from chat interfaces toward multi-agent coordination and decision support in high-stakes domains. International trade makes this distinction observable: dialogue is grounded into concrete tariff and trade-share terms, validation checks their feasibility against physical and policy constraints, and settlement determines the volume of trade actually delivered under network and capacity limitations. Our evaluation follows this complete chain through delayed macroeconomic outcomes, rather than stopping at the bargaining table.

Existing bilateral, multi-issue, and Diplomacy-style benchmarks evaluate payoff division, preference inference, communication, or strategic interaction \cite{lewis2017deal,chan2024negotiationtom,kwon2025astra,meta2022human,guan2024richelieu}. Because their endpoints are typically an agreement or a final match outcome, they do not directly test whether natural-language commitments can be grounded into executable terms and yield tangible value after execution. This gap is nontrivial: a mutually acceptable textual deal may violate binding constraints, overstate available supply, or inadvertently harm the broader economic network once settled.

Three concrete failure modes motivate our benchmark design. First, natural-language dialogue must reliably map to structured policy terms; otherwise, cooperative language remains purely rhetorical and never translates into an actionable intervention. Second, even a valid bilateral commitment can under-deliver when exporter production capacity or third-party import demand binds the actual flow of goods. Third, a term that reduces one country's local target gap (the fraction of a trade target still unmet) can simultaneously lower aggregate Economic Health Index (EHI) by introducing trade-balance costs, inflationary pressure, or unresolved supply gaps in other sectors. A useful benchmark must surface all three failure modes while remaining sufficiently controlled for systematic paired comparison.

\begin{figure}[t!]
    \centering
    \includegraphics[width=0.9\linewidth]{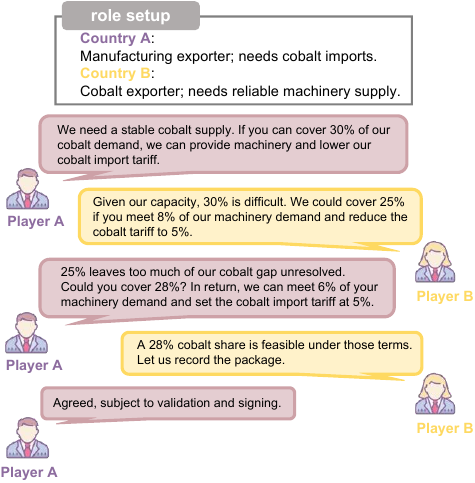}
    \caption{Illustrative negotiation in TradeWar. Two countries bargain over reciprocal cobalt, machinery, and tariff terms; their dialogue-level agreement remains provisional until grounding, validation, and bilateral signing.}
    \Description{Country A, a machinery exporter seeking cobalt, negotiates with Country B, a cobalt exporter seeking machinery. They agree on reciprocal supply shares and an import tariff, subject to grounding, validation, and bilateral signing.}
    \label{fig:negotiation_example}
\end{figure}

To address these gaps, we introduce \textbf{TradeWar}, a reproducible multi-country benchmark in which negotiated tariff and trade-share terms become executable policy interventions (Figure~\ref{fig:negotiation_example}). Multilateral coupling, trade-graph constraints, and capacity-limited settlement surface cases that agreement-only metrics miss: a \emph{superficial agreement trap}, where consensus does not become a valid or delivered commitment, and an \emph{execution-binding failure}, where executable terms nevertheless produce weak downstream value. 

We also introduce \textbf{TACTIC} (Trade-aware Adaptive Consolidation and Theory-of-mind for International Commerce), an agent that adapts during use while keeping its base LLM fixed and maintaining a compact model of each opponent \cite{yao2022react,shinn2023reflexion,madaan2023self}. By combining experience memory, opponent profiles, bounded planning, and structured reflection, TACTIC iteratively updates its negotiation guidance using both the terms signed and the settlement outcomes realized. Its design explicitly decouples strategic adaptation from parametric model updates, enabling robust improvement across episodes without catastrophic forgetting or costly retraining.

Our contributions are threefold:
\begin{itemize}
    \item \textbf{An execution-grounded benchmark:} TradeWar evaluates natural-language agreements through the full pipeline of validation, settlement, and delayed macroeconomic feedback, revealing failures invisible to agreement-rate metrics.
    \item \textbf{An outcome-driven agent:} TACTIC adapts persistent strategic guidance from execution outcomes without modifying base LLM weights, enabling cross-episode learning while preserving general conversational capability.
    \item \textbf{Empirical validation:} Through cross-play, sequential adaptation, stress scenarios, and ablations, we show that agreement rate is a poor proxy for post-execution value and that settlement-grounded adaptation improves beyond CoT prompting and in-episode memory alone.
\end{itemize}

%The remainder of this paper is organized as follows. Section~\ref{sec:related} reviews related work on LLM-based negotiation, economic simulation, and multi-agent adaptation. Section~\ref{sec:benchmark} formally defines the TradeWar benchmark, including the grounding procedure, feasibility validation, and settlement dynamics. Section~\ref{sec:agent} details the TACTIC architecture, its memory and reflection components, and its theory-of-mind opponent modeling. Section~\ref{sec:experiments} presents our experimental setup, main results, ablation studies, and robustness analysis under technology and energy shocks. Finally, Section~\ref{sec:conclusion} concludes with a discussion of limitations, broader implications, and directions for future work.

% =========================================================================
% Section 2: Related Work
% =========================================================================
\section{Related Work}

\subsection{Negotiation Environments and Benchmarks}

Negotiation and strategic interaction have been studied in classical bargaining theory, automated negotiation platforms, and recent LLM-based environments \cite{osborne1990bargaining,davidson2024evaluating,liu2025promediate,mangla2025negotiationgym}. Classical automated negotiation covers protocol design, concession tactics, opponent modeling, multi-issue agenda setting, agent competitions, and platform-based evaluation \cite{kraus2001strategic,jennings2001automated,faratin1998negotiation,fatima2004agenda,jonker2017automated,Baarslag2016ASO,lin2014genius,baarslag2014decoupling}.

Recent environments vary in party count, action format, and evaluation endpoint. Buyer-seller bargaining tasks split a fixed surplus \cite{lewis2017deal}; multi-issue resource allocation games require preference inference and trade-offs \cite{noh2024llms,abdelnabi2024llmdeliberation,chan2024negotiationtom,kwon2025astra,liu2025promediate}; economic games add pricing, private information, and strategic exchange \cite{shapira2024glee,guo2024economics,zhu2025automated}; and Diplomacy-style environments test alliance formation and trust over long horizons \cite{calhamer1974invention,meta2022human,guan2024richelieu,xu2025dipllm,duffy2026democratizing}.

Recent agent-market benchmarks extend this spectrum beyond one-shot bargaining. AgenticPay studies feasible buyer--seller transactions; Cattle Trade and SidConArena combine bargaining with auctions, resource allocation, or production; and CoffeeBench and Magentic Marketplace model supply-chain or marketplace transactions over time \cite{liu2026agenticpay,muller2026cattle,feng2026sidconarena,sugiura2026coffeebench,bansal2025magentic}. These environments primarily evaluate transaction feasibility, bargaining behavior, game score, profit, or market welfare. TradeWar instead focuses on policy-mediated negotiation among six heterogeneous countries: signed tariff/share commitments alter a coupled trade graph, affect third-party access, pass through capacity-constrained quarterly settlement, and change a multi-objective macroeconomic state. This design tests whether LLM agents can ground language into valid policy, reason about network spillovers, and adapt to delayed outcomes and shocks across repeated episodes. Table~\ref{tab:benchmark_comparison} summarizes the resulting differences in evaluation scope.

\begin{table*}[t]
\centering
\caption{Comparison of negotiation and strategic-interaction benchmarks. Symbols: \checkmark{} = supported, $\circ$ = partial, \textemdash{} = absent. TradeWar combines executable commitments with delayed post-execution feedback.}
\label{tab:benchmark_comparison}
\small
\setlength{\tabcolsep}{10pt}
\renewcommand{\arraystretch}{1.2}
\begin{tabular*}{\textwidth}{@{\extracolsep{\fill}}lcccc@{}}
\toprule
\textbf{Benchmark family} &
\textbf{Multi-party} &
\textbf{Executable} &
\makecell[c]{\textbf{Delayed}\\\textbf{post-exec.}\\\textbf{feedback}} &
\textbf{Primary endpoint} \\
\midrule
Bilateral bargaining~\cite{lewis2017deal} &
\textemdash & \textemdash & \textemdash &
Deal / immediate utility \\
Multi-issue negotiation~\cite{chan2024negotiationtom,kwon2025astra,abdelnabi2024llmdeliberation} &
\textemdash & \textemdash & \textemdash &
Agreement quality \\
Economic games / agent markets~\cite{shapira2024glee,liu2026agenticpay,sugiura2026coffeebench,feng2026sidconarena,bansal2025magentic} &
$\circ$ & $\circ$ & $\circ$ &
Payoff / profit / welfare \\
Diplomacy-style~\cite{meta2022human,guan2024richelieu,xu2025dipllm,duffy2026democratizing} &
\checkmark & \textemdash & \textemdash &
Match / board outcome \\
\textbf{TradeWar (ours)} &
\checkmark & \checkmark & \checkmark &
\textbf{Post-execution EHI} \\
\bottomrule
\end{tabular*}
\end{table*}

Some economic games and agent-market environments include binding transactions, repeated interaction, or delayed returns, so Table~\ref{tab:benchmark_comparison} marks their multi-party structure, executable effects, and delayed feedback as partial. TradeWar combines these properties with policy-level commitments, third-party trade-graph effects, delayed multi-quarter settlement, and macroeconomic feedback.

This endpoint gap is especially relevant for LLM-based agents, which can make offers, infer preferences, and adapt in natural-language bargaining or multi-party negotiation settings \cite{lewis2017deal,xia2024measuring,fu2023improvinglanguage,zhan2024let,liu2026opponent,davidson2024evaluating,priya2025genteel,konishi2026preference,liu2025promediate}. Evaluation nevertheless remains largely dialogue-level, emphasizing deal completion, persuasive quality, or immediate utility \cite{lewis2017deal,fu2023improvinglanguage,davidson2024evaluating}. Empirical studies report unstable behavior, limited strategic depth, and inconsistent gains from larger models \cite{bianchi2024well,schneider2024negotiating,kwon2024arellms,shah2025llmrationalis,hakimov2026price}.

Diplomacy-style settings provide a multi-country comparison point because they involve shifting alliances and long-term consequences. Their commitments remain informal: a promise can be broken without formal validation or a measurable economic effect (Table~\ref{tab:benchmark_comparison}). TradeWar instead grounds agreements as executable policy terms and evaluates their delayed post-execution value.

\subsection{Self-Improvement and Continual Adaptation}

At inference time, chain-of-thought prompting can elicit stepwise deliberation without updating model parameters \cite{wei2022chain,kojima2022large}. Related work also examines how language agents update behavior across interactions. ReAct, Reflexion, and Self-Refine use reflection over past actions to improve performance \cite{yao2022react,shinn2023reflexion,madaan2023self}. Surveys describe a move from static prompting toward self-improving systems \cite{dong2024survey,lin2026position,fang2025comprehensive}. Similar approaches have been applied to Diplomacy, negotiation, and opponent modeling \cite{guan2024richelieu,fu2023improvinglanguage,liu2026opponent,yuan2024self,goel2025lifelong,mangla2025negotiationgym}.

Feedback in these systems is usually direct, such as a win, score, or deal indicator. Adaptation from delayed, execution-mediated feedback remains less studied \cite{dong2024survey,lin2026position}. In TradeWar, the value of an agreement is observed only after settlement, where its effect may be mixed with other transactions and external shocks. This delayed and confounded signal makes credit assignment more difficult and provides the feedback used by TACTIC's cross-episode updates.

TradeWar brings these dimensions together at the policy and macroeconomic levels: multi-party interaction, executable commitments, and delayed post-execution feedback.

% =========================================================================
% Section 3: Problem Formulation and the TradeWar Environment 
% =========================================================================
\section{TradeWar: A Long-Horizon Economic Negotiation Environment}

\begin{figure*}[t!]
\centering
\includegraphics[width=0.85\textwidth]{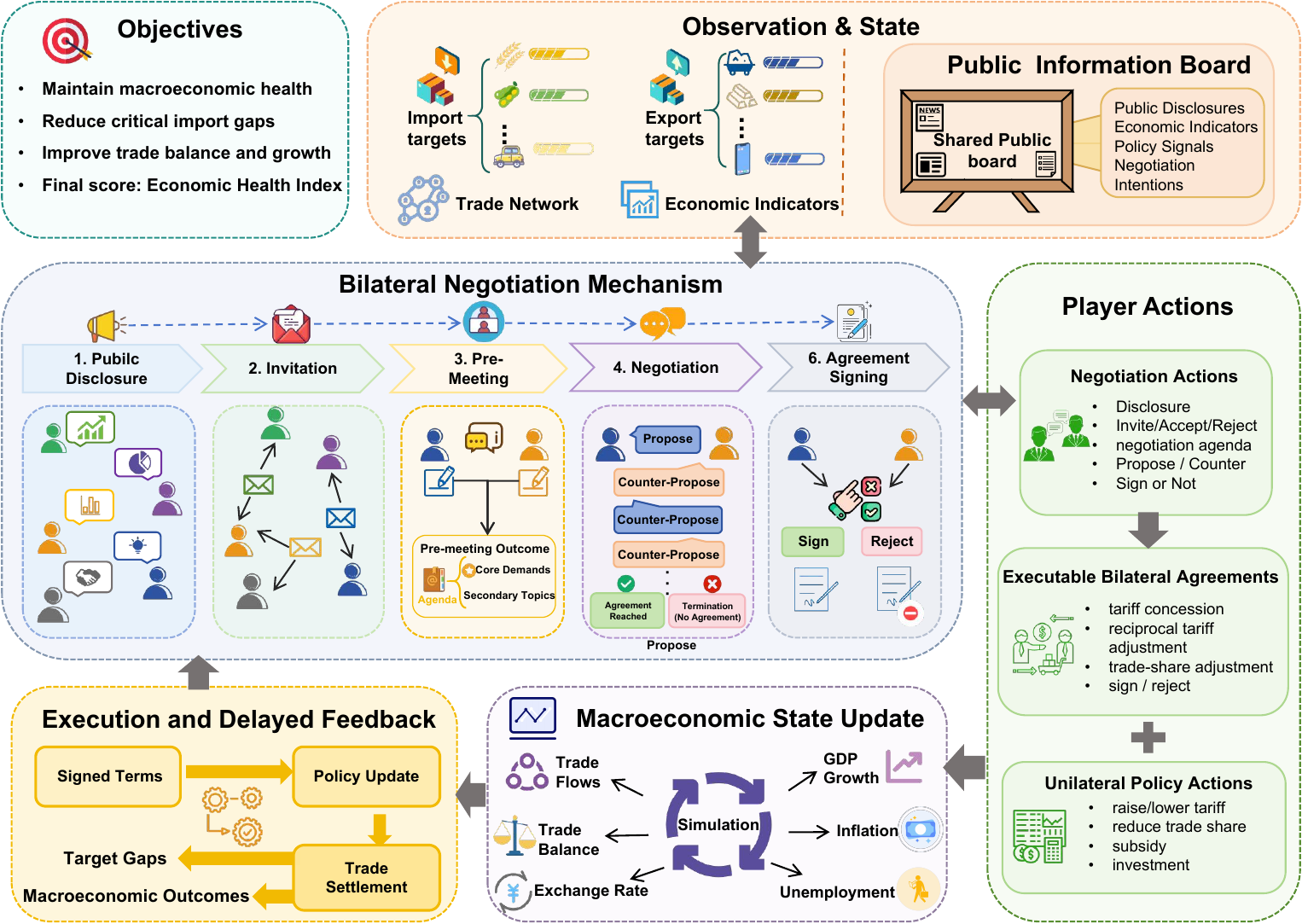}
\caption{Overview of the TradeWar environment. Signed language terms are grounded into policy updates, executed through settlement, and returned as delayed EHI feedback.}
\Description{A pipeline-style overview of the TradeWar environment showing countries acting under partial observability, negotiating in natural language, forming executable bilateral agreements or unilateral policies, grounding signed terms into state changes, propagating them through trade settlement, and receiving delayed macroeconomic feedback measured by EHI.}
\label{fig:environment_design}
\end{figure*}

TradeWar implements a negotiation--execution loop. Negotiation affects environment state only through signed and validated policy terms. Capacity-constrained settlement executes those terms, and the resulting economic effects are observed later as feedback. Figure~\ref{fig:environment_design} summarizes this causal chain:
\begin{equation*}
\text{negotiation} \rightarrow \text{execution} \rightarrow \text{economic transition} \rightarrow \text{delayed feedback}.
\end{equation*}

At the environment level, the Economic Health Index (EHI) aggregates the effects of commitments on stylized macroeconomic conditions using fixed, pre-specified weights (Appendix~\ref{app:implementation}).

\subsection{Design Motivation and Principles}
TradeWar is a controlled agent benchmark rather than a predictive economic simulator, consistent with the use of stylized agent-based environments for controlled comparison \cite{tesfatsion2006handbook,de2014agent}. Trade provides a suitable testbed because tariffs, access shares, and supply targets can be expressed in language and checked mechanically after signing. The paper specifies the evaluation target and state-transition mechanics needed to interpret the benchmark. Three design principles govern the environment: state changes require validated and mutually signed terms; value is assessed after capacity-constrained settlement; and transitions, shocks, and validation are reproducible under fixed seeds.

\subsection{Evaluation Target and State Space}
Six heterogeneous countries trade products $\mathcal{K}$ over multiple years. Country $i$ receives observation $o_t^i=(o_{t,\mathrm{priv}}^i,b_t)$, where the private component contains domestic constraints, inventory pressure, and unmet import/export targets, while the public board $b_t$ exposes tariffs and shared economic indicators. Counterpart urgency, minimum acceptable terms (reservation limits), and private targets remain hidden and must be inferred through repeated interaction.

The endpoint EHI is a fixed benchmark score that aggregates normalized gross domestic product (GDP) growth, inflation, unemployment, trade balance, and critical-supply completion. These components create competing pressures: increasing imports may close a supply gap while worsening balance or inflation, and protecting domestic production may preserve fiscal resources while restricting needed inputs. EHI provides a common multi-objective score for controlled comparison; it is neither an estimate of welfare nor policy advice.

\subsection{Action Space and Game Formulation}
Each country acts on partial information and optimizes its own EHI through dialogue, bilateral agreements, and unilateral policies. During an active negotiation window, a country may invite a counterpart, exchange offers and counteroffers, accept or reject a proposal, and decide whether to sign it. Executable terms adjust importer-side tariffs or bilateral trade shares, while unilateral actions alter the acting country's own import barriers, trade-edge exposure, or domestic industrial support.

These choices create strategic conflict. Protection may relieve domestic pressure but increase import costs or shortages, while reallocating access toward one supplier reduces the opportunities available to others. Affected countries therefore have an incentive to revise their partners, demands, or concessions in later negotiation windows. Formally, TradeWar is a finite-horizon, general-sum partially observable stochastic game \cite{shoham2008multiagent}
$\langle\mathcal{N},\mathcal{S},\{\mathcal{A}_i\},\{\mathcal{O}_i\},P,\{r_i\},\gamma\rangle$.
Given joint action $\mathbf{a}_t$, exogenous shocks $\xi_t$ enter the quarterly transition
$s_{t+1}\sim P(s_{t+1}\mid s_t,\mathbf{a}_t,\xi_t)$.
For policies $\pi_1,\ldots,\pi_N$, long-horizon performance is
\begin{equation}
J_i(\pi_1,\ldots,\pi_N)=
\mathbb{E}\!\left[\sum_{t=0}^{T-1}\gamma^t r_i(s_t,\mathbf{a}_t,s_{t+1})\right].
\label{eq:posg-objective}
\end{equation}
Here, $T$ denotes the number of quarterly transitions; in our experiments, $H=5$ years corresponds to $T=20$ quarters. The step-wise reward $r_i$ represents the incremental change in the agent's Economic Health Index ($\Delta \text{EHI}$). Under an undiscounted setting ($\gamma=1$), $J_i$ captures cumulative EHI improvement; because initial conditions are fixed within matched comparisons, this objective is aligned with the episode-final EHI used as the primary evaluation metric in our experiments. This formulation makes the benchmark general-sum: countries share an interest in stability but compete over access, supply, and trade balance.

\subsection{Negotiation and Executable Agreement}
Negotiation opens once every four quarters and proceeds through disclosure, invitation, agenda exchange, bargaining, and signing. A signed proposal is normalized into tariff and trade-share terms, checked against party and graph constraints, and then applied. For a trade-share term, the granted partner's access is set to $q_{t+1,k}^{j\rightarrow i}=q'$. When the importer still has other active suppliers, access $q'$ granted to partner $j$ proportionally reduces residual access for every $h\neq j$:
\begin{equation}
q_{t+1,k}^{h\rightarrow i}
=q_{t,k}^{h\rightarrow i}
\frac{1-q'}{\sum_{m\neq j}q_{t,k}^{m\rightarrow i}}.
\label{eq:share-reallocation}
\end{equation}
For example, granting one partner a 30\% share allocates the remaining 70\% proportionally among other suppliers. A bilateral concession therefore affects other trade edges.

Validation makes the update mechanical: invalid updates and values outside the feasible share range are rejected before any state change. Grounding uses a fixed agreement schema containing parties, product, importer/exporter direction, term type, magnitude, and duration. A term is accepted only when the parties and directions are consistent, the referenced product edge exists, the policy type is supported, and the resulting shares remain feasible. Invalid or unparseable terms are recorded without changing state. Conversational consensus can therefore occur without an environment-level agreement.

\subsection{Economic Transition and Delayed Feedback}
A signed agreement creates a requested trade flow, while settlement determines the amount delivered. At quarter end, settlement first converts importer demand $D_{t,k}^{i}$ and agreed share into requested flow
$\widehat{x}_{t,k}^{j\rightarrow i}=q_{t,k}^{j\rightarrow i}D_{t,k}^{i}\mathbb{I}_{t,k}^{j\rightarrow i}$,
where $\mathbb{I}$ indicates edge availability. Here, $\widehat{x}$ denotes requested flow, $x$ realized flow, $K$ exporter capacity, and $\lambda>0$ an unnormalized priority weight; the denominator below normalizes weighted requests. If aggregate requests exceed exporter capacity $K_{t,k}^{j}$, settlement allocates realized flow as
\begin{equation}
x_{t,k}^{j\rightarrow i}=
\begin{cases}
\widehat{x}_{t,k}^{j\rightarrow i},
& \sum_r\widehat{x}_{t,k}^{j\rightarrow r}\le K_{t,k}^{j},\\[3pt]
K_{t,k}^{j}
\dfrac{\lambda_{t,k}^{j\rightarrow i}\widehat{x}_{t,k}^{j\rightarrow i}}
{\sum_r\lambda_{t,k}^{j\rightarrow r}\widehat{x}_{t,k}^{j\rightarrow r}},
& \text{otherwise}.
\end{cases}
\label{eq:capacity-rationing}
\end{equation}
Thus even a valid agreement can under-deliver. Realized trade determines balance $B_t^i$ and contributes to the stylized macroeconomic update
\begin{align}
G_{t+1}^{i} &= G_t^{i}(1+\Delta g_t^{i}),\\
\Delta g_t^{i}
&=\alpha_0+\alpha_B\widetilde{B}_t^{i}+\alpha_U U_t^{i}
+\alpha_R R_t^{i}-\alpha_M M_t^{i}+\epsilon_t^{i},
\end{align}
where $G$ is GDP, $\widetilde{B}$ normalized trade balance, $U$ unilateral support, $R$ fiscal contribution, $M$ import burden, and $\epsilon$ a disturbance term. Fixed coefficients govern these benchmark mechanics rather than estimated elasticities. Because the resulting signal appears after settlement and may combine several transactions, agents must distinguish conversational agreement, executable commitment, and realized value.

\subsection{Scenario Perturbations and Ethical Considerations}
Exogenous events $\xi_t$ alter capacity, edge availability, or fiscal conditions, testing whether agents revise strategies that succeeded under the baseline. These shocks are controlled perturbations, not forecasts of real events. In the deterministic stress tests used in this paper, both shocks are introduced in the first quarter (Q1) of the starting year and persist thereafter. The technology-controls scenario disables a designated high-technology export edge and imposes a prohibitive bilateral tariff barrier, whereas the energy-crisis scenario halves the designated energy exporter's capacity together with the corresponding maximum-capacity, target-volume, and public export-cap parameters. Shock schedules are fixed under a seed so agents face matched conditions. Country labels and shock names are stylized scenario labels; the dynamics are synthetic and non-predictive, so TradeWar evaluates multi-agent behavior rather than forecasting policy.

% =========================================================================
% Section 4: Architecture
% =========================================================================
\section{Architecture}

TACTIC is organized around TradeWar's three failure points. Structured planning and deterministic signing checks reduce grounding failures; execution-scored memory and reflection connect signed terms to realized settlement outcomes; and opponent profiles with fallback planning support decisions under partial observability and network spillovers. These modules update guidance stored outside the fixed base LLM \cite{guan2024richelieu,dong2024survey}.

\subsection{Overview}

\begin{figure}[t!]
    \centering
    \includegraphics[width=\linewidth]{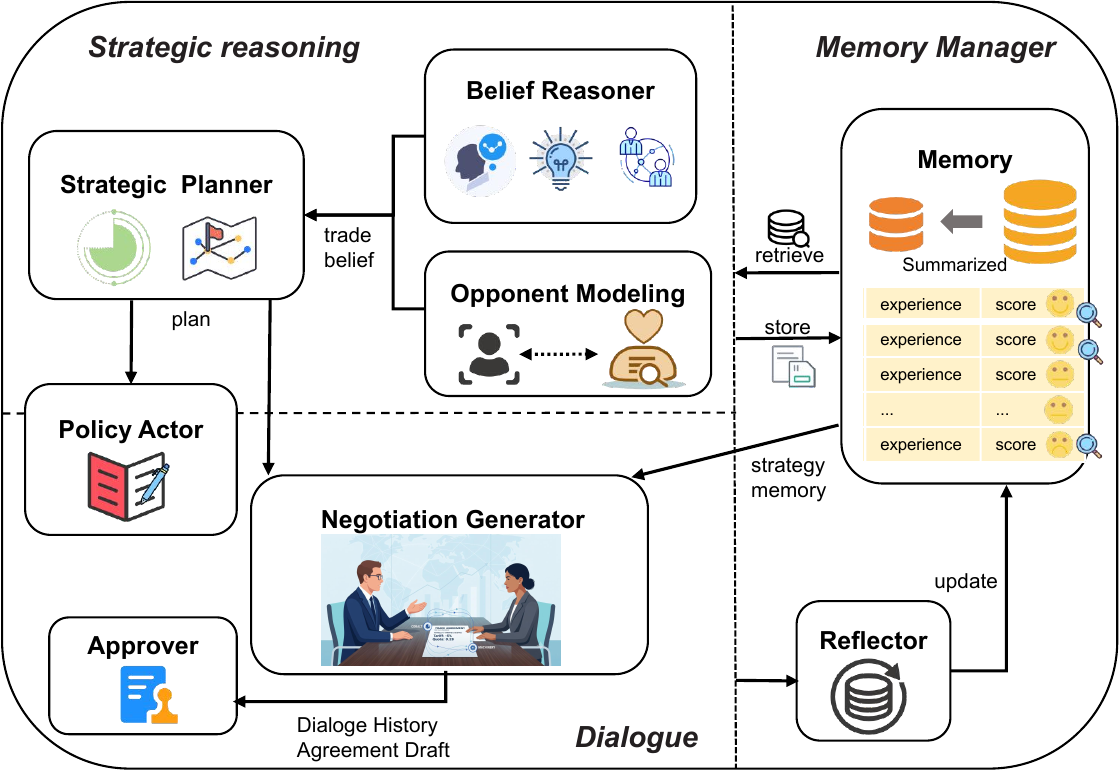}
\caption{TACTIC's three-stage execution-feedback loop and four core modules.}
\Description{A modular architecture diagram for TACTIC with three stages: strategic reasoning, dialogue and action, and post-settlement reflection with memory update. Four core modules---experience memory, structured reflection, opponent modeling, and strategy planning---connect observations and plans to signed agreements, delayed outcomes, and updated strategy state.}
\label{fig:tactic_architecture}
\end{figure}

Figure~\ref{fig:tactic_architecture} organizes the pipeline into three stages. Strategic reasoning first combines observation $o_t^i$, retrieved experience, and a compact opponent profile into a bounded intent plan. Dialogue and action then turn that plan into messages and candidate agreements, with feasibility checks before signing. Finally, settlement feedback triggers reflection and updates persistent state for the next decision. Strategic intent is computed separately from surface dialogue, as in prior modular agent designs \cite{huang2024understanding,li2025review}. Heuristic rules handle memory scoring, profile updates, and signing checks; the LLM handles planning, dialogue generation, and qualitative diagnosis. This split keeps state updates auditable while allowing flexible language generation.

\subsection{Experience Memory}

Each memory record stores counterpart, proposed terms, signing status, and post-settlement feedback. An outcome score combines execution validity, target-gap reduction, and macroeconomic effect. Retrieval first filters by partner and product relevance and then ranks the remaining records by this score, prioritizing commitments with realized effects over recent dialogue alone. The memory manager supplies compact summaries to the planner, limiting context growth while preserving the connection between a proposed commitment and its realized consequence. The resulting memory represents execution-scored experience in addition to transcript history \cite{park2023generative,wang2025memento}.

\subsection{Structured Reflection}

Reflection converts delayed feedback into reusable guidance \cite{shinn2023reflexion,madaan2023self}. Its prompt jointly considers the intent plan, parsed signed terms, realized flows, target-gap changes, and macroeconomic outcome. The output assigns a \emph{superficial agreement trap} or \emph{execution-binding failure}, records counterpart notes, and extracts a concise lesson such as changing the agenda, partner, or term magnitude. Numeric memory and profile updates remain heuristic rather than being assigned directly by the LLM.

\subsection{Opponent Modeling}

The opponent model maintains a compact profile $\mathcal{P}_t^i$ of trust, predictability, and concession tendency rather than a full latent mental state \cite{Baarslag2016ASO,chan2024negotiationtom,liu2026opponent}. Profile updates are deterministic functions of observable signing behavior, changes in stated demands across turns, execution success, and settlement outcomes. A compact profile summary is exposed to planning and dialogue, where it bounds packages that exceed a counterpart's likely acceptance range or delivery capacity. A signed term that under-delivers therefore lowers inferred reliability regardless of cooperative dialogue.

\subsection{Strategy Planning}

The planner maps economic goals into a structured intent $\chi_t^i$ containing a focal product, primary demand, concession bound, and fallback \cite{faratin1998negotiation,fatima2004agenda,jonker2017automated,fu2023improvinglanguage}. If an edge is unavailable, it can redirect the proposal to a feasible partner or product and attach a reciprocal tariff concession. Dialogue generation must remain consistent with this intent, while deterministic signing checks reject unsupported terms before submission. This bounded planner does not solve for equilibrium \cite{hao2026game}; it narrows generation toward signable, opponent-aware packages.

\subsection{Supporting Action Stages}

The four modules condition, but do not replace, the stages that interact with the environment. A context builder summarizes the current private observation, public board, retrieved records, and opponent profile for the planner. The negotiation generator then produces both a natural-language response and, when appropriate, a structured proposal containing supported tariff or share terms. Keeping strategic intent separate from surface realization allows the same plan to support counteroffers without regenerating priorities each turn.

Before signing, an algorithmic checker verifies edge existence, party direction, term type, value bounds, and consistency with the active intent. The checker prevents a linguistically valid but unsupported proposal from being treated as an executed action; it does not decide whether a strategically weak but feasible agreement should be accepted. When negotiation is unavailable or insufficient, a policy actor may instead choose a supported unilateral action from the same belief and plan. These supporting stages connect persistent strategy state and language generation to environment execution, which accounts for the additional components in Figure~\ref{fig:tactic_architecture}.

\subsection{Outcome-Driven Cross-Episode Adaptation}

Execution outcomes update persistent memory, opponent profiles, and strategy guidance across episodes---\emph{cross-episode adaptation}---without changing model parameters \cite{lin2026position,fang2025comprehensive}. Let $\psi_t^i$ be strategy guidance and $\omega_t^i$ summarize executability, target-gap change, and economic outcomes. After settlement,
\begin{equation}
    (\mathcal{M}_{t+1}^i,\mathcal{P}_{t+1}^i,\psi_{t+1}^i)
    =
    \mathcal{U}\!\left(\mathcal{M}_t^i,\mathcal{P}_t^i,\psi_t^i,\omega_t^i\right).
\label{eq:strategy-update}
\end{equation}
The update $\mathcal{U}$ combines heuristic memory scoring and profile revision with LLM-based reflection and planning. Failed delivery can lower counterpart reliability, downgrade the associated experience, and shift later plans toward fallback partners. These updates persist across episodes, allowing future decisions to retrieve execution-validated experience in addition to within-episode dialogue. TACTIC carries forward only $\mathcal{M}^i$, $\mathcal{P}^i$, and $\psi^i$; the base LLM, prompt templates, validation rules, and environment dynamics remain fixed.

% =========================================================================
% Section 5: Experiments
% =========================================================================
\section{Experiments}

Four research questions (RQs) ask whether TradeWar distinguishes agents beyond agreement statistics (\textbf{RQ1}), whether outcome-based updates improve performance across episodes (\textbf{RQ2}), how agents respond to controlled shocks (\textbf{RQ3}), and how the TACTIC modules contribute cumulatively (\textbf{RQ4}). Code, frozen configurations, and logging scripts are available at
\textcolor{blue}{\textit{\url{https://github.com/kxyzc1992/TradeWar-Benchmark}}}.
Across all protocols, the study comprises approximately 500 full episodes. Each episode is one $H=5$-year run with $T=20$ quarterly transitions, sparse negotiation windows, agreement validation, settlement, and delayed feedback.

\subsection{Experimental Setup}

\paragraph{Environment.}
Experiments use six anonymized countries, a horizon of $H=5$ years (20 quarterly updates), and three-versus-three (3v3) agent assignments. One agent type controls three countries and its opponent controls the remainder. Except for RQ2's fixed sequential assignment, every seed is paired with an episode in which the two agent types exchange country assignments, controlling for country asymmetry. We evaluate a calibrated no-shock baseline plus \texttt{technology\_controls} and \texttt{energy\_crisis}.

\paragraph{Cross-play protocol (RQ1/RQ4).}
RQ1 uses cross-play to test whether agent differences persist across opponents; RQ4 reuses the same opponent pool for module ablations. Each of five agent types is evaluated against the other four, with five episodes per ordered matchup (100 baseline games). Country assignments are balanced across both directions of each pair. Figure~\ref{fig:rq1_heatmap} reports mean final EHI ($\times 100$) and sample standard deviation for the row agent; Table~\ref{tab:baseline_comparison} aggregates EHI and secondary metrics over the same pool.

\paragraph{Sequential protocol (RQ2).}
RQ2 tests whether outcome-driven updates help across repeated episodes. Against a fixed PromptLLM and fixed assignment, MemoryLLM, TACTIC-static, and full TACTIC each run 20 sequential episodes over three independent seeds (180 total). For each seed, the compared methods face the same episode order and initial conditions. Figure~\ref{fig:rq2_curve} reports three-episode rolling final EHI with across-seed bands; early and late values are means over episodes 1--3 and 18--20. TACTIC-static retains CoT prompting, memory, opponent modeling, planning, and reflection within an episode but carries no outcome-driven memory, profile, or strategy updates across episodes. MemoryLLM uses CoT with working, episodic, and semantic interaction memory within an episode, but its stores are reset between sequential episodes. These controls separate persistent settlement-grounded adaptation from stepwise deliberation, interaction history, and module count alone.

\paragraph{Perturbation protocol (RQ3).}
RQ3 tests robustness under controlled perturbations. TACTIC, MemoryLLM, and PromptLLM are tested under baseline and both shocks using 12 paired-mirror episodes per condition (the same six seeds under two reversed role assignments; 108 total). Relative drops use each agent's matched no-shock run from this stress test rather than Table~\ref{tab:baseline_comparison}.

\paragraph{Evaluation Metrics.}\label{sec:evaluation}
The primary outcome is mean final EHI ($\times 100$ in reported tables and figures), the pre-specified weighted aggregate of GDP growth, unemployment, inflation, trade balance, and separate import/export target-completion scores defined in Appendix~\ref{app:implementation}. Cumulative GDP growth and product-target completion are reported as diagnostic metrics. Agreement rate counts dialogue-level accepted proposals before post-settlement value is observed and is reported separately as a dialogue comparator, not an outcome metric.

\paragraph{Baselines and LLM backend.}
We build two LLM-based baselines. \textbf{PromptLLM} receives the current private observation and public state together with a zero-shot CoT instruction to reason step by step about target gaps, executable constraints, counterpart incentives, and likely settlement consequences before returning only schema-constrained JSON \cite{wei2022chain,kojima2022large}. \textbf{MemoryLLM} uses the same CoT prompt and additionally retrieves within-episode interaction memory; it does not retain settlement-grounded updates across episodes. RuleBased uses macro-pressure and target-gap heuristics to propose and sign small executable packages and choose unilateral policies, while Random samples supported invitations, proposals, policies, and signing decisions. RQ2 also includes TACTIC-static, which keeps TACTIC's within-episode memory, opponent modeling, planning, and reflection but resets persistent state across episodes.

At each action-generation stage, all LLM agents receive the same schema-constrained CoT instruction, use GPT-4o-mini with \texttt{max\_tokens}=1000 and temperature $0.3$, and emit actions through the same structured interface. Thus, the comparisons vary the additional state and update mechanisms rather than the availability of stepwise deliberation. Held-out conditions use separate memory stores. Every run logs its scenario, horizon, negotiation schedule, seed, assignment, generation settings, and memory protocol.

\paragraph{Reproducibility controls.}
Environment transitions, validation, and scheduled shocks are deterministic given a seed; stochasticity comes primarily from model generation. Reversed role assignments address heterogeneous countries, and separate memory stores prevent experience from leaking across methods or conditions. We report episode-level means and sample standard deviations because EHI and the secondary metrics are terminal episode outcomes; the present sample budgets do not support strong asymptotic significance claims.

\subsection{Results}

\begin{figure}[t]
\centering
\includegraphics[width=0.92\linewidth]{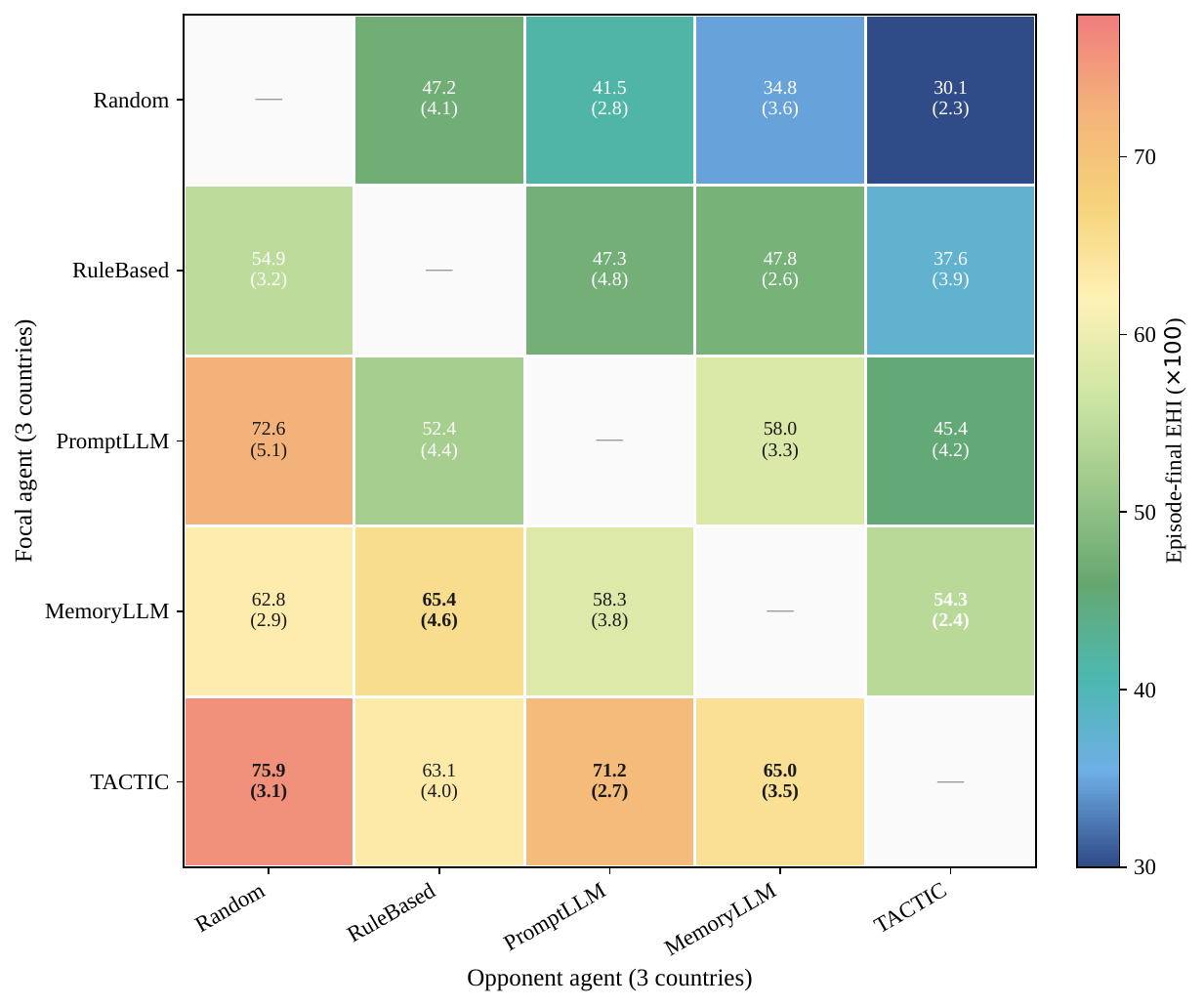}
\caption{Pairwise baseline cross-play over 100 games. Rows denote focal agents and columns denote opponents; each off-diagonal cell reports mean final EHI ($\times 100$) with sample standard deviation over five episodes. Warmer colors indicate higher focal-agent EHI; same-agent matchups are excluded.}
\Description{A square heatmap of pairwise cross-play results among five agent types. Rows denote the focal agent and columns denote the opponent. Each off-diagonal cell reports mean final EHI with standard deviation. The pattern shows stronger performance for TACTIC in most opponent matchups, while same-agent matchups are omitted on the diagonal.}
\label{fig:rq1_heatmap}
\end{figure}

\textbf{TACTIC vs. Baselines under the Baseline Scenario.}
Table~\ref{tab:baseline_comparison} shows the central gap between agreement and post-execution value: despite using CoT prompting, PromptLLM reaches 75\% agreement but only 57.1 EHI. MemoryLLM, which adds within-episode interaction memory, reaches 60.2, while TACTIC reaches 68.8. Thus, CoT prompting and interaction memory alone do not ensure that an accepted proposal produces post-settlement value. Figure~\ref{fig:rq1_heatmap} resolves these aggregates by opponent. TACTIC achieves the highest focal-agent EHI in three of four matchups---against Random, PromptLLM, and MemoryLLM---whereas MemoryLLM slightly exceeds TACTIC against RuleBased (65.4 versus 63.1). This pairwise variation supports TACTIC's aggregate advantage without implying a strict total ranking (RQ1).

\begin{table}[t]
\centering
\caption{Baseline cross-play results. EHI is the row mean of Figure~\ref{fig:rq1_heatmap}.}
\label{tab:baseline_comparison}
\begingroup
\small
\setlength{\tabcolsep}{0pt}
\renewcommand{\arraystretch}{1.08}
\begin{tabular*}{\linewidth}{@{\extracolsep{\fill}}lcccc@{}}
\toprule
\textbf{Agent} & \textbf{Agree. (\%)} & \textbf{EHI} & \textbf{GDP (\%)} & \textbf{Target (\%)} \\
\midrule
Random & 15 & 38.4 & -0.6 & 9.2 \\
RuleBased & 35 & 46.9 & 0.5 & 28.4 \\
PromptLLM & 75 & 57.1 & 1.3 & 43.6 \\
MemoryLLM & 80 & 60.2 & 2.1 & 54.8 \\
\textbf{TACTIC} & \textbf{85} & \textbf{68.8} & \textbf{4.0} & \textbf{71.3} \\
\bottomrule
\end{tabular*}
\endgroup
\end{table}

\begin{figure}[t]
\centering
\includegraphics[width=\linewidth]{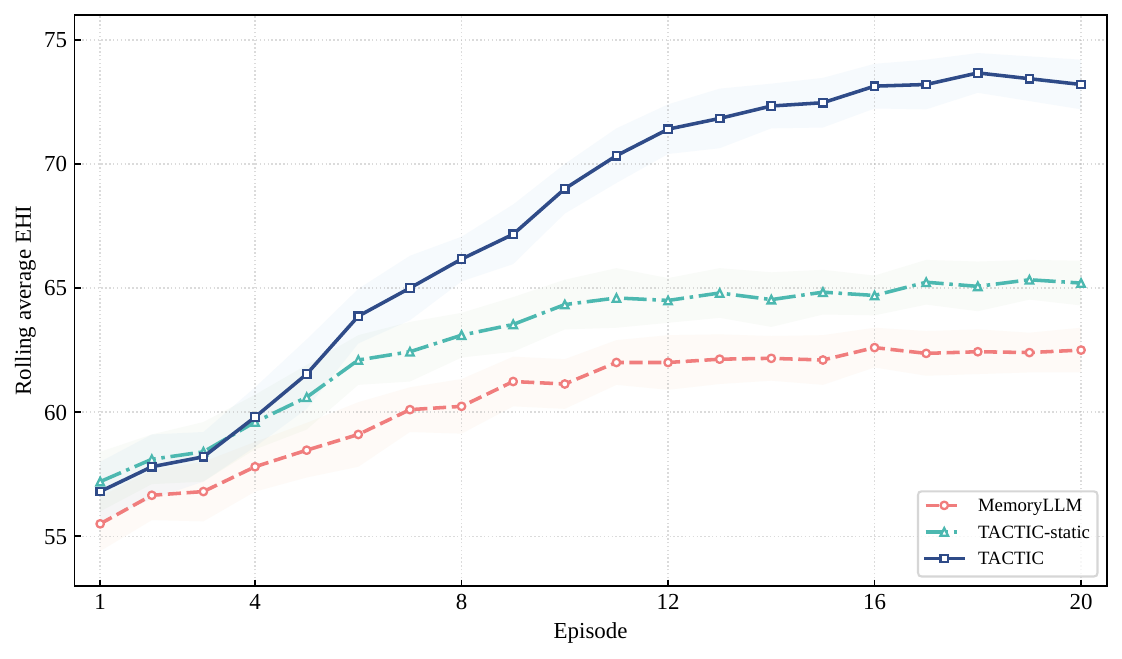}
\caption{Three-seed rolling EHI over 20 sequential episodes against fixed PromptLLM. TACTIC-static disables cross-episode updates; bands show across-seed variation.}
\Description{A line chart over 20 sequential episodes comparing MemoryLLM, TACTIC-static, and full TACTIC on rolling final EHI. Full TACTIC climbs the most and finishes highest, TACTIC-static improves more modestly, and MemoryLLM plateaus lower. Shaded bands indicate variation across seeds.}
\label{fig:rq2_curve}
\end{figure}

\textbf{Outcome-Driven Adaptation over Repeated Episodes.}
The first-three-episode means of TACTIC-static and full TACTIC are nearly identical (58.4 vs.\ 58.2), consistent with their identical CoT instruction and within-episode modules. By episodes 18--20, MemoryLLM and TACTIC-static plateau at 62.5 and 65.2, whereas full TACTIC reaches 73.2 (Figure~\ref{fig:rq2_curve}). Because TACTIC-static already includes CoT, memory, opponent modeling, planning, and reflection, the 8.0-point gap more directly isolates the benefit of persistent settlement-grounded updates (RQ2).
The within-episode modules account for similar initial performance, while only full TACTIC carries execution outcomes into later strategy.

\begin{figure}[t]
\centering
\includegraphics[width=\linewidth]{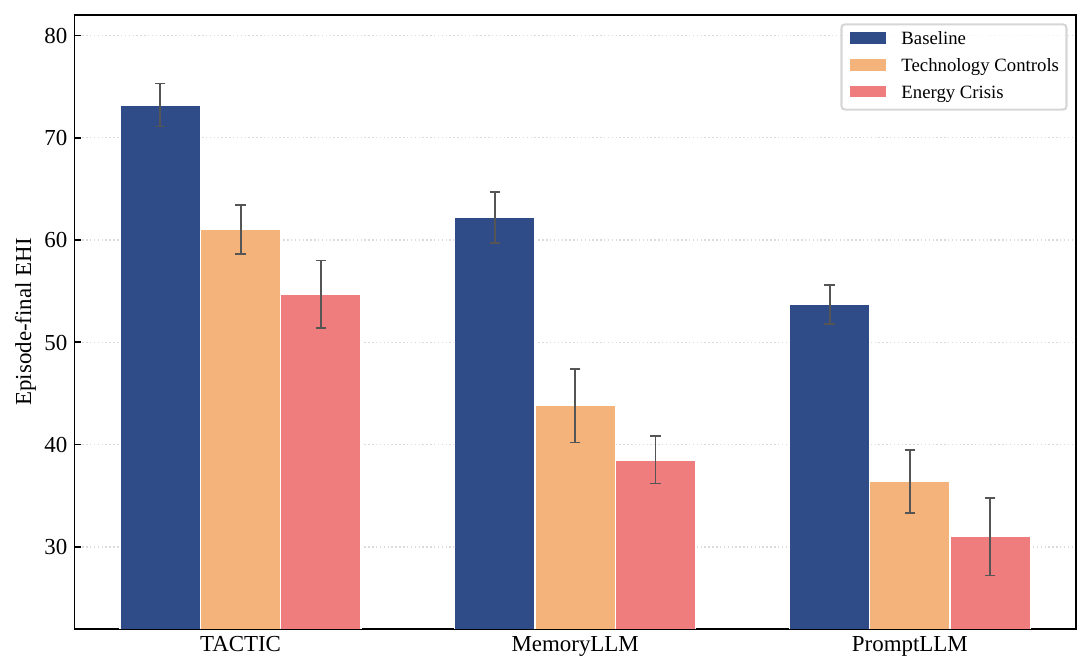}
\caption{Episode-final EHI under baseline and shock scenarios. Each condition uses 12 episodes (six seeds under reversed role assignments); error bars show sample standard deviation.}
\Description{A comparative robustness plot showing episode-final EHI for TACTIC, MemoryLLM, and PromptLLM under baseline, technology-controls, and energy-crisis conditions. TACTIC remains highest across conditions and exhibits smaller relative declines under shocks than the simpler baselines. Error bars show sample variation.}
\label{fig:rq3_robustness}
\end{figure}

\textbf{TACTIC under Scenario Perturbations.}
The technology-controls scenario reduces TACTIC by 16.7\%, versus 29.6\% for MemoryLLM and 32.2\% for PromptLLM; under the energy-crisis scenario, the corresponding drops are 25.3\%, 38.1\%, and 42.3\%, respectively (Figure~\ref{fig:rq3_robustness}). TACTIC still loses absolute performance, but degrades least in both shocks (RQ3).
These results establish relative robustness rather than invariance to shocks. Every agent declines more under the energy-crisis scenario than under the technology-controls scenario, but CoT prompting and within-episode memory alone show larger losses than settlement-grounded adaptation.

\textbf{Ablation Study of TACTIC Modules.}
EHI rises at each cumulative step in Table~\ref{tab:ablation}. The first row is the zero-module endpoint of the cumulative TACTIC ablation: it retains the shared CoT instruction and TACTIC action-stage scaffold but supplies no adaptive module state. We use it as the starting point of the cumulative ablation rather than as an additional external baseline. Adding reflection to memory, opponent modeling, and planning raises EHI from 63.9 to 68.8, GDP growth by 1.3 points, and target completion by 15.1 points. This final incremental comparison isolates reflection's contribution, whereas the effects of earlier modules may depend on their addition order (RQ4). The table supports the assembled pipeline but does not provide order-independent attribution for every component.

\begin{table}[t]
\centering
\caption{Cumulative ablation under the baseline cross-play protocol.}
\label{tab:ablation}
\begingroup
\small
\setlength{\tabcolsep}{0pt}
\renewcommand{\arraystretch}{1.08}
\begin{tabular*}{\linewidth}{@{\extracolsep{\fill}}lccc@{}}
\toprule
\textbf{Configuration} & \textbf{EHI} & \textbf{GDP (\%)} & \textbf{Target (\%)} \\
\midrule
No adaptive modules & 51.6 & 0.8 & 32.1 \\
+ Experience Memory & 55.3 & 1.4 & 38.7 \\
+ Opponent Modeling & 59.0 & 1.9 & 47.4 \\
\midrule
\makecell[l]{\textbf{TACTIC w/o Reflection}\\(memory + opponent + planning)} & 63.9 & 2.7 & 56.2 \\
\makecell[l]{\textbf{Full TACTIC}\\(+ structured reflection)} & 68.8 & 4.0 & 71.3 \\
\bottomrule
\end{tabular*}
\endgroup
\end{table}

\textbf{Qualitative failure diagnosis.}
The aggregate results correspond to two trace patterns detailed in Appendix~\ref{app:cases}. In a superficial agreement trap, dialogue converges on broad market-access language, but parsing, graph constraints, or signing checks prevent a state change. In an execution-binding failure, valid terms yield little target-gap improvement because of capacity rationing or a poorly chosen edge. Successful adaptations change the next executable target or package, for example by switching product or partner and narrowing the reciprocal concession. The traces illustrate why agreement rate can over-rank fluent agents and why persistent adaptation uses post-settlement feedback.

\section{Discussion}

TradeWar exposes a hierarchy obscured by agreement-only evaluation. A proposal may be accepted yet fail validation; a valid signed term may under-deliver during settlement; and delivered trade may still produce weak aggregate value. Agreement rate therefore captures only an early stage of the pipeline. PromptLLM's high agreement rate but lower final EHI illustrates why execution validity, realized delivery, and post-execution value should be evaluated separately.

The sequential results further isolate the adaptation signal. TACTIC-static retains the same within-episode modules but cannot reuse settlement outcomes; its lower late-stage performance suggests that improvement depends on connecting proposed terms to their eventual effects. Persistent updates revise partner, product, or concession choices after capacity limits and network spillovers are observed. TACTIC thus adapts executable interventions rather than only the persuasiveness of later messages. This evidence remains specific to TradeWar's controlled synthetic dynamics, but demonstrates the value of testing whether agents connect language, feasible action, and delayed consequences.

\section{Conclusion}

TradeWar evaluates natural-language agreements through grounding, settlement, and delayed economic feedback. Its results show that agreement rate does not fully characterize post-execution value. TACTIC adapts persistent guidance through memory, opponent modeling, bounded planning, and reflection while keeping model parameters fixed. It improves more across episodes than MemoryLLM and TACTIC-static, and degrades less under shocks than the CoT-prompted PromptLLM and MemoryLLM baselines; cumulative ablations further support the assembled pipeline. These findings motivate evaluating strategic LLM agents by the executability and realized value of their commitments in addition to dialogue-level agreement.

\section{Limitations and Future Work}

TradeWar is a controlled benchmark and is not intended to model real diplomacy or estimate economic welfare. Its synthetic countries, fixed transition rules, and pre-specified EHI weights support reproducible comparison while limiting external validity. Introducing the environment and TACTIC together also creates a risk of co-adaptation. Delayed feedback and partial observability favor memory and reflection, so TACTIC's advantage may not transfer to simpler bargaining domains with immediate feedback. Agent rankings may also change under different EHI weights. TACTIC should therefore be interpreted as a strong baseline for studying execution-grounded feedback, with transfer to other negotiation settings left for further evaluation.

Future work should test whether execution-grounded adaptation transfers to other domains, evaluate multiple LLM backbones, and explore stronger negotiation baselines. Extending the environment to renegotiable commitments and richer private information will further clarify the boundary between conversational fluency and strategic execution.

\section*{Ethics and Privacy Statement}

This work studies LLM agents in a stylized and anonymized trade-negotiation environment that is not intended for real-world policy recommendation. Methods that improve strategic persuasion or adaptive bargaining could nevertheless be misused in high-stakes economic or diplomatic contexts. We therefore present TradeWar as a controlled research testbed and state its bounded, non-predictive scope throughout the paper.

\begin{acks}
This work was supported by the National Natural Science Foundation of China (No.~62406034), the State Key Lab of General Artificial Intelligence at Peking University.
\end{acks}

\bibliographystyle{ACM-Reference-Format}
\bibliography{refs}

@inproceedings{bianchi2024well,
  title={How Well Can LLMs Negotiate? NegotiationArena Platform and Analysis},
  author={Bianchi, Federico and Chia, Patrick John and Yuksekgonul, Mert and Tagliabue, Jacopo and Jurafsky, Dan and Zou, James},
  booktitle={International Conference on Machine Learning},
  pages={3935--3951},
  year={2024},
  organization={PMLR}
}

@inproceedings{xia2024measuring,
  title={Measuring bargaining abilities of llms: A benchmark and a buyer-enhancement method},
  author={Xia, Tian and He, Zhiwei and Ren, Tong and Miao, Yibo and Zhang, Zhuosheng and Yang, Yang and Wang, Rui},
  booktitle={Findings of the Association for Computational Linguistics: ACL 2024},
  pages={3579--3602},
  year={2024}
}

@inproceedings{chan2024negotiationtom,
  title={Negotiationtom: A benchmark for stress-testing machine theory of mind on negotiation surrounding},
  author={Chan, Chunkit and Jiayang, Cheng and Yim, Yauwai and Deng, Zheye and Fan, Wei and Li, Haoran and Liu, Xin and Zhang, Hongming and Wang, Weiqi and Song, Yangqiu},
  booktitle={Findings of the Association for Computational Linguistics: EMNLP 2024},
  pages={4211--4241},
  year={2024}
}

@article{shapira2024glee,
  title={Glee: A unified framework and benchmark for language-based economic environments},
  author={Shapira, Eilam and Madmon, Omer and Reinman, Itamar and Amouyal, Samuel Joseph and Reichart, Roi and Tennenholtz, Moshe},
  journal={arXiv preprint arXiv:2410.05254},
  year={2024}
}

@article{liu2025promediate,
  title={ProMediate: A Socio-cognitive framework for evaluating proactive agents in multi-party negotiation},
  author={Liu, Ziyi and Sarrafzadeh, Bahar and Zhou, Pei and Yang, Longqi and Zhao, Jieyu and Sharma, Ashish},
  journal={arXiv preprint arXiv:2510.25224},
  year={2025}
}

@article{mangla2025negotiationgym,
  title={NegotiationGym: Self-Optimizing Agents in a Multi-Agent Social Simulation Environment},
  author={Mangla, Shashank and Hokamp, Chris and Boylan, Jack and Ghalandari, Demian Gholipour and Jauhari, Yuuv and Cassidy, Lauren and Duffy, Oisin},
  journal={arXiv preprint arXiv:2510.04368},
  year={2025}
}

@book{shoham2008multiagent,
  title={Multiagent systems: Algorithmic, game-theoretic, and logical foundations},
  author={Shoham, Yoav and Leyton-Brown, Kevin},
  year={2008},
  publisher={Cambridge University Press}
}

@article{yao2022react,
  title={React: Synergizing reasoning and acting in language models},
  author={Yao, Shunyu and Zhao, Jeffrey and Yu, Dian and Du, Nan and Shafran, Izhak and Narasimhan, Karthik and Cao, Yuan},
  journal={arXiv preprint arXiv:2210.03629},
  year={2022}
}

@article{shinn2023reflexion,
  title={Reflexion: Language agents with verbal reinforcement learning},
  author={Shinn, Noah and Cassano, Federico and Gopinath, Ashwin and Narasimhan, Karthik and Yao, Shunyu},
  journal={Advances in neural information processing systems},
  volume={36},
  pages={8634--8652},
  year={2023}
}

@inproceedings{lewis2017deal,
  title={Deal or no deal? end-to-end learning of negotiation dialogues},
  author={Lewis, Mike and Yarats, Denis and Dauphin, Yann and Parikh, Devi and Batra, Dhruv},
  booktitle={Proceedings of the 2017 Conference on Empirical Methods in Natural Language Processing},
  pages={2443--2453},
  year={2017}
}

@inproceedings{Baarslag2016ASO,
  title={A Survey of Opponent Modeling Techniques in Automated Negotiation},
  author={Tim Baarslag and Mark J. C. Hendrikx and Koen V. Hindriks and Catholijn M. Jonker},
  booktitle={Adaptive Agents and Multi-Agent Systems},
  year={2016},
  url={https://api.semanticscholar.org/CorpusID:14243697}
}

@book{tesfatsion2006handbook,
  title={Handbook of computational economics: agent-based computational economics},
  author={Tesfatsion, Leigh and Judd, Kenneth L},
  volume={2},
  year={2006},
  publisher={Elsevier}
}

@article{de2014agent,
  title={Agent-based models},
  author={De Marchi, Scott and Page, Scott E},
  journal={Annual Review of political science},
  volume={17},
  number={1},
  pages={1--20},
  year={2014},
  publisher={Annual Reviews}
}

@inproceedings{priya2025genteel,
  title={Genteel-negotiator: Llm-enhanced mixture-of-expert-based reinforcement learning approach for polite negotiation dialogue},
  author={Priya, Priyanshu and Chigrupaatii, Rishikant and Firdaus, Mauajama and Ekbal, Asif},
  booktitle={Proceedings of the aaai conference on artificial intelligence},
  volume={39},
  number={23},
  pages={25010--25018},
  year={2025}
}

@misc{calhamer1974invention,
  author       = {Calhamer, Allan},
  title        = {The Invention of Diplomacy},
  year         = {1974},
  url          = {https://diplomacyzines.co.uk/strategy-tactics/articles-by-alan-b-calhamer/the-invention-of-diplomacy/},
  note         = {Accessed: 2026-07-06}
}

@article{meta2022human,
  title={Human-level play in the game of Diplomacy by combining language models with strategic reasoning},
  author={Bakhtin, Anton and Brown, Noam and Dinan, Emily and Farina, Gabriele and Flaherty, Colin and Fried, Daniel and Goff, Andrew and Gray, Jonathan and Hu, Hengyuan and others},
  journal={Science},
  volume={378},
  number={6624},
  pages={1067--1074},
  year={2022},
  publisher={American Association for the Advancement of Science}
}

@inproceedings{guan2024richelieu,
  title={Richelieu: Self-Evolving {LLM}-Based Agents for {AI} Diplomacy},
  author={Guan, Zhenyu and Kong, Xiangyu and Wang, Yizhou and Zhong, Fangwei},
  booktitle={Advances in Neural Information Processing Systems 37},
  year={2024},
  pages={123471--123497},
  doi={10.52202/079017-3925},
  publisher={Neural Information Processing Systems Foundation, Inc.}
}

@article{xu2025dipllm,
  title={DipLLM: Fine-tuning LLM for strategic decision-making in diplomacy},
  author={Xu, Kaixuan and Chai, Jiajun and Li, Sicheng and Fu, Yuqian and Zhu, Yuanheng and Zhao, Dongbin},
  journal={arXiv preprint arXiv:2506.09655},
  year={2025}
}

@inproceedings{duffy2026democratizing,
  title={Democratizing diplomacy: A harness for evaluating any large language model on full-press diplomacy},
  author={Duffy, Alexander and Paech, Samuel J and Shastri, Ishana and Karpinski, Elizabeth and Alloui-Cros, Baptiste and Marques, Tyler and Olson, Matthew Lyle},
  booktitle={Proceedings of the AAAI Conference on Artificial Intelligence},
  volume={40},
  number={44},
  pages={37350--37359},
  year={2026}
}

@article{goel2025lifelong,
  title={Lifelong sotopia: Evaluating social intelligence of language agents over lifelong social interactions},
  author={Goel, Hitesh and Zhu, Hao},
  journal={arXiv preprint arXiv:2506.12666},
  year={2025}
}

@article{liu2026agenticpay,
  title={AgenticPay: A Multi-Agent LLM Negotiation System for Buyer-Seller Transactions},
  author={Liu, Xianyang and Gu, Shangding and Song, Dawn},
  journal={arXiv preprint arXiv:2602.06008},
  year={2026}
}

@article{sugiura2026coffeebench,
  title={CoffeeBench: Benchmarking Long-Horizon LLM Agents in Heterogeneous Multi-Agent Economies},
  author={Sugiura, Issa and Hattori, Daichi and Araragi, Kazuo and Ogawa, Keita and Onose, Shota and Makino, Taro and Usuki, Teppei and Ishida, Takashi},
  journal={arXiv preprint arXiv:2606.16613},
  year={2026}
}

@article{muller2026cattle,
  title={Cattle Trade: A Multi-Agent Benchmark for LLM Bluffing, Bidding, and Bargaining},
  author={M{\"u}ller, Robert and M{\"u}ller, Clemens},
  journal={arXiv preprint arXiv:2605.14537},
  year={2026}
}

@article{feng2026sidconarena,
  title={SidConArena: An Environment Evaluating Agents in Open-Ended, Positive-Sum Bargaining Game},
  author={Feng, Yeqi and Chen, Yuxin and He, Tianxing},
  journal={arXiv preprint arXiv:2606.27397},
  year={2026}
}

@article{bansal2025magentic,
  title={Magentic Marketplace: An Open-Source Environment for Studying Agentic Markets},
  author={Bansal, Gagan and Hua, Wenyue and Huang, Zezhou and Fourney, Adam and Swearngin, Amanda and Epperson, Will and Payne, Tyler and Hofman, Jake M and Lucier, Brendan and Singh, Chinmay and others},
  journal={arXiv preprint arXiv:2510.25779},
  year={2025}
}

@misc{liu2026opponent,
title={Opponent Simulation as Inference-time Scaling for Self-improving Agent: Case Study of Repeated Negotiations},
author={Xiangyu Liu and Di Wang and Zhe Feng and Aranyak Mehta},
year={2026},
url={https://openreview.net/forum?id=ywXgFYs44d}
}

@article{yuan2024self,
  title={Self-rewarding language models},
  author={Yuan, Weizhe and Pang, Richard Yuanzhe and Cho, Kyunghyun and Li, Xian and Sukhbaatar, Sainbayar and Xu, Jing and Weston, Jason},
  journal={arXiv preprint arXiv:2401.10020},
  year={2024}
}

@article{huang2024understanding,
  title={Understanding the planning of llm agents: A survey},
  author={Huang, Xu and Liu, Weiwen and Chen, Xiaolong and Wang, Xingmei and Wang, Hao and Lian, Defu and Wang, Yasheng and Tang, Ruiming and Chen, Enhong},
  journal={arXiv preprint arXiv:2402.02716},
  year={2024}
}

@article{madaan2023self,
  title={Self-refine: Iterative refinement with self-feedback},
  author={Madaan, Aman and Tandon, Niket and Gupta, Prakhar and Hallinan, Skyler and Gao, Luyu and Wiegreffe, Sarah and Alon, Uri and Dziri, Nouha and Prabhumoye, Shrimai and Yang, Yiming and others},
  journal={Advances in neural information processing systems},
  volume={36},
  pages={46534--46594},
  year={2023}
}

@inproceedings{park2023generative,
  title={Generative agents: Interactive simulacra of human behavior},
  author={Park, Joon Sung and O'Brien, Joseph and Cai, Carrie Jun and Morris, Meredith Ringel and Liang, Percy and Bernstein, Michael S},
  booktitle={Proceedings of the 36th annual acm symposium on user interface software and technology},
  pages={1--22},
  year={2023}
}

@INPROCEEDINGS{zhan2024let,
  title={Let’s negotiate! a survey of negotiation dialogue systems},
  author={Zhan, Haolan and Wang, Yufei and Li, Zhuang and Feng, Tao and Hua, Yuncheng and Sharma, Suraj and Qu, Lizhen and Azad, Zhaleh Semnani and Zukerman, Ingrid and Haf, Reza},
  booktitle={Findings of the Association for Computational Linguistics: EACL 2024},
  pages={2019--2031},
  year={2024}
}

@article{wang2025memento,
  title={Memento-II: Learning by Stateful Reflective Memory},
  author={Wang, Jun},
  journal={arXiv preprint arXiv:2512.22716},
  year={2025}
}

@article{fu2023improvinglanguage,
  title={Improving language model negotiation with self-play and in-context learning from ai feedback},
  author={Fu, Yao and Peng, Hao and Khot, Tushar and Lapata, Mirella},
  journal={arXiv preprint arXiv:2305.10142},
  year={2023}
}

@inproceedings{kwon2025astra,
  title={ASTRA: A Negotiation Agent with Adaptive and Strategic Reasoning via Tool-integrated Action for Dynamic Offer Optimization},
  author={Kwon, Deuksin and Hae, Jiwon and Clift, Emma and Shamsoddini, Daniel and Gratch, Jonathan and Lucas, Gale},
  booktitle={Proceedings of the 2025 Conference on Empirical Methods in Natural Language Processing},
  pages={16217--16238},
  year={2025}
}

@inproceedings{konishi2026preference,
  title={Preference Estimation via Opponent Modeling in Multi-Agent Negotiation},
  author={Konishi, Yuta and Yamamoto, Kento and Sonomoto, Eisuke and Takeda, Rikuho and Furukawa, Ryo and Muraki, Yusuke and Shimizu, Takafumi and Fukumura, Kazuma and Kanemoto, Yuya and Ito, Takayuki and others},
  booktitle={Findings of the Association for Computational Linguistics: ACL 2026},
  pages={28049--28058},
  year={2026}
}

@article{noh2024llms,
  title={Llms with personalities in multi-issue negotiation games},
  author={Noh, Sean and Chang, Ho-Chun Herbert},
  journal={arXiv preprint arXiv:2405.05248},
  year={2024}
}

@inproceedings{li2025review,
  title={A review of prominent paradigms for llm-based agents: Tool use, planning (including rag), and feedback learning},
  author={Li, Xinzhe},
  booktitle={Proceedings of the 31st international conference on computational linguistics},
  pages={9760--9779},
  year={2025}
}

@article{lin2026position,
  title={Position: Agentic Evolution is the Path to Evolving LLMs},
  author={Lin, Minhua and Lu, Hanqing and Shi, Zhan and He, Bing and Mao, Rui and Zhang, Zhiwei and Wu, Zongyu and Tang, Xianfeng and Liu, Hui and Dai, Zhenwei and others},
  journal={arXiv preprint arXiv:2602.00359},
  year={2026}
}

@article{fang2025comprehensive,
  title={A comprehensive survey of self-evolving ai agents: A new paradigm bridging foundation models and lifelong agentic systems},
  author={Fang, Jinyuan and Peng, Yanwen and Zhang, Xi and Wang, Yingxu and Yi, Xinhao and Zhang, Guibin and Xu, Yi and Wu, Bin and Liu, Siwei and Li, Zihao and others},
  journal={arXiv preprint arXiv:2508.07407},
  year={2025}
}

@article{dong2024survey,
  title={A survey on llm inference-time self-improvement},
  author={Dong, Xiangjue and Teleki, Maria and Caverlee, James},
  journal={arXiv preprint arXiv:2412.14352},
  year={2024}
}

@inproceedings{
abdelnabi2024llmdeliberation,
title={{LLM}-Deliberation: Evaluating {LLM}s with Interactive Multi-Agent Negotiation Game},
author={Sahar Abdelnabi and Amr Gomaa and Sarath Sivaprasad and Lea Sch{\"o}nherr and Mario Fritz},
booktitle={ICLR 2024 Workshop on Large Language Model (LLM) Agents},
year={2024},
url={https://openreview.net/forum?id=eE1WHn6qlk}
}

@article{hao2026game,
  title={Game-Theoretic Lens on LLM-based Multi-Agent Systems},
  author={Hao, Jianing and Ding, Han and Xu, Yuanjian and Sun, Tianze and Chen, Ran and Zhang, Wanbo and Zhang, Guang and Li, Siguang},
  journal={arXiv preprint arXiv:2601.15047},
  year={2026}
}

@article{zhu2025automated,
  title={The automated but risky game: Modeling and benchmarking agent-to-agent negotiations and transactions in consumer markets},
  author={Zhu, Shenzhe and Sun, Jiao and Nian, Yi and South, Tobin and Pentland, Alex and Pei, Jiaxin},
  journal={arXiv preprint arXiv:2506.00073},
  year={2025}
}

@inproceedings{hakimov2026price,
  title={The price of thought: A multilingual analysis of reasoning, performance, and cost of negotiation in large language models},
  author={Hakimov, Sherzod and Bernard, Roland and Leiber, Tim and Osswald, Karl and Richert, Kristina and Yang, Ruilin and Bernardi, Raffaella and Schlangen, David},
  booktitle={Findings of the Association for Computational Linguistics: EACL 2026},
  pages={529--570},
  year={2026}
}

@article{guo2024economics,
  title={Economics arena for large language models},
  author={Guo, Shangmin and Bu, Haoran and Wang, Haochuan and Ren, Yi and Sui, Dianbo and Shang, Yuming and Lu, Siting},
  journal={arXiv preprint arXiv:2401.01735},
  year={2024}
}

@inproceedings{schneider2024negotiating,
  title={Negotiating with llms: Prompt hacks, skill gaps, and reasoning deficits},
  author={Schneider, Johannes and Haag, Steffi and Kruse, Leona Chandra},
  booktitle={International conference on computer-human interaction research and applications},
  pages={238--259},
  year={2024},
  organization={Springer}
}

@inproceedings{kwon2024arellms,
  title={Are llms effective negotiators? systematic evaluation of the multifaceted capabilities of llms in negotiation dialogues},
  author={Kwon, Deuksin and Weiss, Emily and Kulshrestha, Tara and Chawla, Kushal and Lucas, Gale and Gratch, Jonathan},
  booktitle={Findings of the Association for Computational Linguistics: EMNLP 2024},
  pages={5391--5413},
  year={2024}
}

@article{shah2025llmrationalis,
  title={LLM Rationalis? Measuring bargaining capabilities of AI negotiators},
  author={Shah, Cheril and Agarwal, Akshit and Garg, Kanak and Heddaya, Mourad},
  journal={arXiv preprint arXiv:2512.13063},
  year={2025}
}

@inproceedings{davidson2024evaluating,
  title={Evaluating language model agency through negotiations},
  author={Davidson, Tim R and Veselovsky, Veniamin and Kosinski, Michal and West, Robert},
  booktitle={International Conference on Learning Representations},
  volume={2024},
  pages={784--809},
  year={2024}
}

@inproceedings{jonker2017automated,
  title={Automated negotiating agents competition (ANAC)},
  author={Jonker, Catholijn and Aydogan, Reyhan and Baarslag, Tim and Fujita, Katsuhide and Ito, Takayuki and Hindriks, Koen},
  booktitle={Proceedings of the AAAI conference on artificial intelligence},
  volume={31},
  number={1},
  year={2017}
}

@article{jennings2001automated,
  title={Automated negotiation: prospects, methods and challenges},
  author={Jennings, Nicholas R and Faratin, Peyman and Lomuscio, Alessio R and Parsons, Simon and Sierra, Carles and Wooldridge, Michael},
  journal={International Journal of Group Decision and Negotiation},
  volume={10},
  number={2},
  pages={199--215},
  year={2001}
}

@article{lin2014genius,
  title={Genius: An integrated environment for supporting the design of generic automated negotiators},
  author={Lin, Raz and Kraus, Sarit and Baarslag, Tim and Tykhonov, Dmytro and Hindriks, Koen and Jonker, Catholijn M},
  journal={Computational Intelligence},
  volume={30},
  number={1},
  pages={48--70},
  year={2014},
  publisher={Wiley Online Library}
}

@incollection{baarslag2014decoupling,
  title={Decoupling negotiating agents to explore the space of negotiation strategies},
  author={Baarslag, Tim and Hindriks, Koen and Hendrikx, Mark and Dirkzwager, Alexander and Jonker, Catholijn},
  booktitle={Novel insights in agent-based complex automated negotiation},
  pages={61--83},
  year={2014},
  publisher={Springer}
}

@article{faratin1998negotiation,
  title={Negotiation decision functions for autonomous agents},
  author={Faratin, Peyman and Sierra, Carles and Jennings, Nick R},
  journal={Robotics and Autonomous Systems},
  volume={24},
  number={3-4},
  pages={159--182},
  year={1998},
  publisher={Elsevier}
}

@article{fatima2004agenda,
  title={An agenda-based framework for multi-issue negotiation},
  author={Fatima, Shaheen S and Wooldridge, Michael and Jennings, Nicholas R},
  journal={Artificial Intelligence},
  volume={152},
  number={1},
  pages={1--45},
  year={2004},
  publisher={Elsevier}
}

@book{kraus2001strategic,
  title={Strategic negotiation in multiagent environments},
  author={Kraus, Sarit},
  year={2001},
  publisher={The MIT press}
}

@book{osborne1990bargaining,
  title={Bargaining and Markets},
  author={Osborne, Martin J. and Rubinstein, Ariel},
  year={1990},
  publisher={Academic Press}
}

@article{wei2022chain,
  title={Chain-of-thought prompting elicits reasoning in large language models},
  author={Wei, Jason and Wang, Xuezhi and Schuurmans, Dale and Bosma, Maarten and Xia, Fei and Chi, Ed and Le, Quoc V and Zhou, Denny and others},
  journal={Advances in neural information processing systems},
  volume={35},
  pages={24824--24837},
  year={2022}
}

@article{kojima2022large,
  title={Large language models are zero-shot reasoners},
  author={Kojima, Takeshi and Gu, Shixiang Shane and Reid, Machel and Matsuo, Yutaka and Iwasawa, Yusuke},
  journal={Advances in neural information processing systems},
  volume={35},
  pages={22199--22213},
  year={2022}
}

% OPTIONAL APPENDICES: after references, with no page limit.
% Remove this whole block, including \clearpage, if there are no appendices.
\clearpage
\appendix

\section{Implementation Details}
\label{app:implementation}

\subsection{Rules of the TradeWar Game}

This section complements Section~3 by providing the operational and implementation-level rules of the environment. It specifies observability constraints, available actions, agreement validation, and the precise computation of economic consequences.

\begin{table}[h]
\centering
\caption{Compact benchmark specification.}
\label{tab:benchmark_spec}
\small
\begin{tabular}{p{0.28\linewidth}p{0.64\linewidth}}
\toprule
\textbf{Component} & \textbf{Specification} \\
\midrule
Agents & Six country agents with heterogeneous private constraints, public indicators, and product-level trade targets. \\
Temporal unit & Quarterly transitions over multi-year episodes; formal negotiation opens once every four quarters. \\
Interaction protocol & Bilateral invitations, agenda exchange, alternating-turn bargaining, and a separate signing stage; bargaining is capped at 50 rounds. \\
Action categories & Natural-language negotiation messages, unilateral policy actions, and executable bilateral tariff/share agreements. \\
Agreement schema & Parties, product, importer/exporter direction, term type, magnitude, and duration. \\
Validation & Party consistency, edge existence, supported term type, directional consistency, and graph/share feasibility. \\
Feedback & Signed terms are applied before settlement; capacity-constrained settlement produces delayed feedback and final EHI ($\times 100$). \\
Stress tests & Deterministic \texttt{technology\_controls} and \texttt{energy\_crisis} scenarios under fixed seeds. \\
\bottomrule
\end{tabular}
\end{table}

\subsubsection{Objectives and Quarterly Rhythm}

Agents operate over a multi-year horizon divided into quarterly steps. Performance is evaluated at the end of the episode using the Economic Health Index (EHI), which rewards strong macroeconomic conditions and the successful completion of trade targets. 

Negotiation windows are sparse, opening once every four quarters. During an active window, agents may disclose information, exchange bilateral invitations, negotiate over multiple rounds, and sign agreements. In non-negotiation quarters, diplomacy is paused; the environment applies existing policies, settles trade, updates macroeconomic indicators, and reveals the delayed consequences of previously executed decisions.

\subsubsection{Observation and Available Actions}

The environment enforces partial observability. Each agent observes its own macroeconomic state, product-level trade status, recent trade summaries, and a shared public board containing global tariff schedules and shared economic signals. An opponent's internal urgency, private priorities, and unfulfilled trade targets remain hidden and must be inferred from negotiation behavior and historical interactions.

Agent actions are strictly categorized into three types:
\begin{itemize}
    \item \textbf{Negotiation actions:} Natural-language messages used to persuade, request concessions, reject terms, or propose agreements.
    \item \textbf{Unilateral policy actions:} Independent adjustments to an agent's own import barriers, trade edge exposures, or domestic industrial subsidies. These do not require mutual consent.
    \item \textbf{Executable bilateral agreements:} Mutually signed proposals that secure new market access or adjust bilateral tariff/share commitments.
\end{itemize}

\subsubsection{Invitation, Bargaining, and Signing Protocol}

Negotiation is available only in Q1 of each year. Agents may issue multiple bilateral invitations during the invitation phase; each accepted invitation opens one bilateral session, subject to the environment's tradable-overlap checks. A short pre-meeting then establishes an agenda and tradable product scope, and formal bargaining proceeds by alternating turns for at most 50 rounds.

A negotiation session is treated as agreed once an actionable proposal is accepted, or when both parties accept within the recent turn window; it breaks immediately on rejection or explicit termination, and unresolved sessions expire at the round cap. Signing is a separate stage entered only after negotiation agreement. The environment canonicalizes the agreed proposal into executable tariff/share terms and asks both parties for a signing decision; only bilateral acceptance yields a signed agreement, and unsigned or non-executable proposals do not change state.

\subsubsection{Agreement Grounding and Validation Rules}

To transition from natural-language dialogue to a state-changing policy, a signed agreement must be grounded into structured terms. A parsed term is only executable if it satisfies all of the following feasibility constraints:
\begin{enumerate}
    \item \textbf{Party consistency:} The term exclusively involves the two negotiating countries and assigns them consistent importer/exporter roles.
    \item \textbf{Edge existence:} The referenced product edge currently exists in the global trade graph.
    \item \textbf{Executable term type:} The term corresponds to a supported policy action (e.g., an importer-side tariff commitment or a bilateral trade-share adjustment).
    \item \textbf{Graph feasibility:} Applying the term preserves all trade-graph constraints, including partner-share normalization ($ \sum q \le 1 $).
\end{enumerate}

Messages that fail parsing or violate any constraint are discarded. If a directional mistake is unambiguous under the current graph, the normalizer may automatically repair it, recording the intervention in the execution trace. 

\subsubsection{Import/Export Targets and Economic Consequences}

Countries are assigned specific volume targets for critical imports (necessary for domestic stability) and exports (necessary for trade balance). At the end of each quarter, realized trade flows are compared against these targets. 

Signed agreements do not guarantee delivery. If global demand for a country's exports exceeds its available capacity, the environment rations supply across partners. A formally executable agreement may therefore underperform when promised supply cannot be realized. Tariff policies also have delayed macroeconomic consequences: higher tariffs may protect domestic industries while exacerbating shortages or inflationary pressure.

\subsubsection{EHI Scoring and Metric Units}

EHI can be computed after each quarterly transition using six normalized scores in $[0,1]$; the primary reported metric is the final-quarter EHI. Let $\operatorname{clip}_{[0,1]}(\cdot)$ denote standard bounding. The macroeconomic scores are defined as:
\begin{align*}
S_g(g) &= \operatorname{clip}_{[0,1]}\left(\frac{g+5}{15}\right),\\
S_{\ell}(\ell) &= 1-\operatorname{clip}_{[0,1]}\left(\frac{\ell}{25}\right),\\
S_{\rho}(\rho) &= 1-\operatorname{clip}_{[0,1]}\left(\frac{|\rho-2|}{6}\right),\\
S_B(B,\widetilde{G}) &= 1-\operatorname{clip}_{[0,1]}\left(\frac{|B/\widetilde{G}|}{0.10}\right).
\end{align*}
Here, $g$, $\ell$, and $\rho$ represent percentage-valued GDP growth, unemployment, and inflation, respectively, while $\widetilde{G}$ is GDP normalized to the trade-value unit used for the balance ratio $B$. 

For export and import target completion, let $\mathcal{K}_\star$ denote the relevant product subset, $v_k$ the realized volume, $\bar v_k$ the target volume, and $w_k$ the product importance weight. We define
\begin{equation*}
S_{\mathrm{target}}(\mathcal{K}_\star)
=
1-\operatorname{clip}_{[0,1]}
\left(
\frac{\sum_{k\in\mathcal{K}_\star} w_k |v_k-\bar v_k|}
{\sum_{k\in\mathcal{K}_\star} w_k \bar v_k+\varepsilon}
\right),
\quad \varepsilon=10^{-8}.
\end{equation*}
We then instantiate separate export and import scores as
\begin{equation*}
S_{\mathrm{exp}} = S_{\mathrm{target}}(\mathcal{K}_{\mathrm{exp}}),
\qquad
S_{\mathrm{imp}} = S_{\mathrm{target}}(\mathcal{K}_{\mathrm{imp}}).
\end{equation*}
The final Economic Health Index is the fixed weighted sum
\begin{equation*}
\mathrm{EHI}
=
0.20\,S_g
+0.15\,S_{\ell}
+0.15\,S_{\rho}
+0.20\,S_B
+0.15\,S_{\mathrm{exp}}
+0.15\,S_{\mathrm{imp}}.
\end{equation*}
These weights are pre-specified benchmark parameters shared across all agents and scenarios. Internally, EHI lies in $[0,1]$; throughout the paper we report $\mathrm{EHI}\times 100$ for readability. Consistent with the main text, EHI is an evaluation aggregate for agent comparison, not an economic welfare function.

\subsubsection{Scenario Perturbations}

Exogenous scenario events ($\xi_t$) are used to stress-test agent robustness. In the reported experiments, perturbations are scheduled deterministically at Q1 of the starting year. Under \texttt{energy\_crisis}, the designated energy exporter's capacity is reduced by a factor of $0.5$; specifically, the affected export product's production capacity, maximum capacity, target volume, and published export cap are all halved. Under \texttt{technology\_controls}, a designated high-technology export edge is disabled and the associated bilateral tariff barrier is set to $1.0$. These fixed benchmark specifications are applied consistently across all runs.

\subsection{Prompt Templates}

For clarity and to facilitate reproduction, we provide the prompt templates for three core TACTIC modules below. At runtime, the placeholders in these templates are instantiated with the corresponding public-board state ($b_t$), private observation ($o_{t,\mathrm{priv}}^i$), and required output-format fields.

\subsubsection{Trade Belief and Intent Planning Prompt}
Before negotiation, TACTIC combines its private state, opponent profile, and retrieved memory into a strategic intent plan ($\chi_t^i$).

\begin{quote}
\small
\textbf{System:} You are the chief trade strategist for [Country\_Name]. Your goal is to improve your country's economic position by supporting growth, limiting inflation and unemployment pressure, maintaining trade balance, and securing critical supply targets.\\
\textbf{Context:}
- Current Quarter: [t]
- Your Private State: [Unfulfilled supply gaps, Inventory levels, Fiscal urgency]
- Opponent Profile ([Opponent\_Name]): [Inferred needs, Past reliability, Concession tendency]
- Retrieved Strategy Memories: [Relevant past experiences from $\mathcal{M}_t^i$]

\textbf{Task:} Generate a Trade Belief and Intent Plan.
1. Identify your most critical import needs and export markets.
2. Infer [Opponent\_Name]'s likely reservation limits based on the trade network.
3. Define your negotiation agenda: What is your primary demand? What are you willing to concede (e.g., lower import tariffs on specific goods)? What is your fallback option if the negotiation fails?
\end{quote}

\subsubsection{Negotiation and Proposal Generation Prompt}
During a bargaining window, the agent uses the intent plan to draft both natural-language dialogue and executable policy terms.

\begin{quote}
\small
\textbf{System:} You are negotiating a bilateral trade agreement with [Opponent\_Name]. Your response should remain consistent with your current Intent Plan: [$\chi_t^i$].\\
\textbf{Context:}
- Recent Dialogue History: [Last N messages]
- Feasibility Constraints: You cannot offer market access you do not possess, nor can you alter tariffs on non-existent trade edges.

\textbf{Task:}
1. Provide a persuasive natural-language response to [Opponent\_Name]'s last message.
2. If appropriate, draft a formal policy proposal. The proposal MUST be a valid JavaScript Object Notation (JSON) array of executable terms (e.g., \{"type": "tariff\_change", "product": "advanced\_tech", "new\_rate": 0.05\}). Do not propose terms that violate the environment's grounding constraints.
\end{quote}

\subsubsection{Post-Execution Reflection Prompt}
After quarterly settlement, TACTIC uses the delayed feedback signal ($\omega_t^i$) to assess the realized effects of past commitments. The reflection prompt records the diagnosis, counterpart notes, and strategy lesson. Numeric opponent scores (e.g., trust) are subsequently updated by heuristic rules, and the resulting guidance is reused in later planning prompts.

\begin{quote}
\small
\textbf{System:} A quarter has passed since you signed the trade agreement with [Opponent\_Name]. Evaluate its actual impact in the environment.\\
\textbf{Context:}
- The Signed Agreement: [Parsed executable terms]
- Your Expectations (Intent Plan): [What you hoped to achieve]
- Delayed Settlement Feedback: [Realized trade flows, Supply shortfalls, Changes in gaps and macro indicators]

\textbf{Task:}
1. Did the agreement succeed after execution, or did it show a \textit{superficial agreement trap} or an \textit{execution-binding failure} (e.g., the counterpart could not deliver the promised supply)?
2. Record qualitative notes on [Opponent\_Name]'s reliability and concession behavior for later use.
3. Write a short strategy lesson to guide future negotiations under similar conditions.
\end{quote}

% Appendix B: designed illustrative cases (grounded in logged run patterns / calibrated
% roles), presented as representative negotiation--execution examples.
% Mechanics constraints reflected in the narratives:
%   - formal negotiation only in Q1 each year; Q2--Q4 are settlement-only
%   - Country_C is a manufacturing exporter of Vehicles/Machinery (not an importer from B)
%   - gap values are target-gap ratios; EHI reported as country EHI x100
% Figures to be supplied by authors:
%   figs/app_case_b1_adaptation.pdf
%   figs/app_case_b2_agreement_outcome.pdf
%   figs/app_case_b3_bounded_commitment.pdf
% Prompt for Fig.~B.1: figs/PROMPT_app_case_b1.txt

\section{Qualitative Analysis of Execution-Grounded Negotiation Dynamics}
\label{app:cases}

This appendix presents three representative negotiation--execution examples grounded in logged run patterns. Following qualitative analyses of prior adaptive strategic agents~\cite{guan2024richelieu}, each case holds strategic pressure approximately fixed and compares a short-sighted or dialogue-optimized path with an execution-aware path. The cases examine successful adaptation, a failure mode, and a role-conditioned strategy choice through changes in the \emph{executable target}, \emph{signed package}, and \emph{post-settlement score}.

Each case includes a dual-track timeline, condensed dialogue, measured economic consequences, and a diagnosis of the observed difference. The dialogue retains only decision-relevant turns. Under the environment calendar, formal negotiation occurs in \textbf{Q1} of each year; Q2--Q4 are settlement-only quarters in which existing policies are applied and delayed consequences become visible. Timeline figures mark the Q1 negotiation window (light) and subsequent settlement quarters (dark), with nodes for bargaining stalls, signing or rejection, settlement feedback, and, for cross-episode adaptation, agenda revision at the \emph{next} Q1 window. Gap values are \emph{target-gap ratios} (the unmet fraction of a product target), and EHI values are country-level scores on the paper's $\times 100$ scale. The cases provide trace-level context for the quantitative evaluation in Section~5.

\begin{figure*}[t]
\centering
\includegraphics[width=0.95\linewidth]{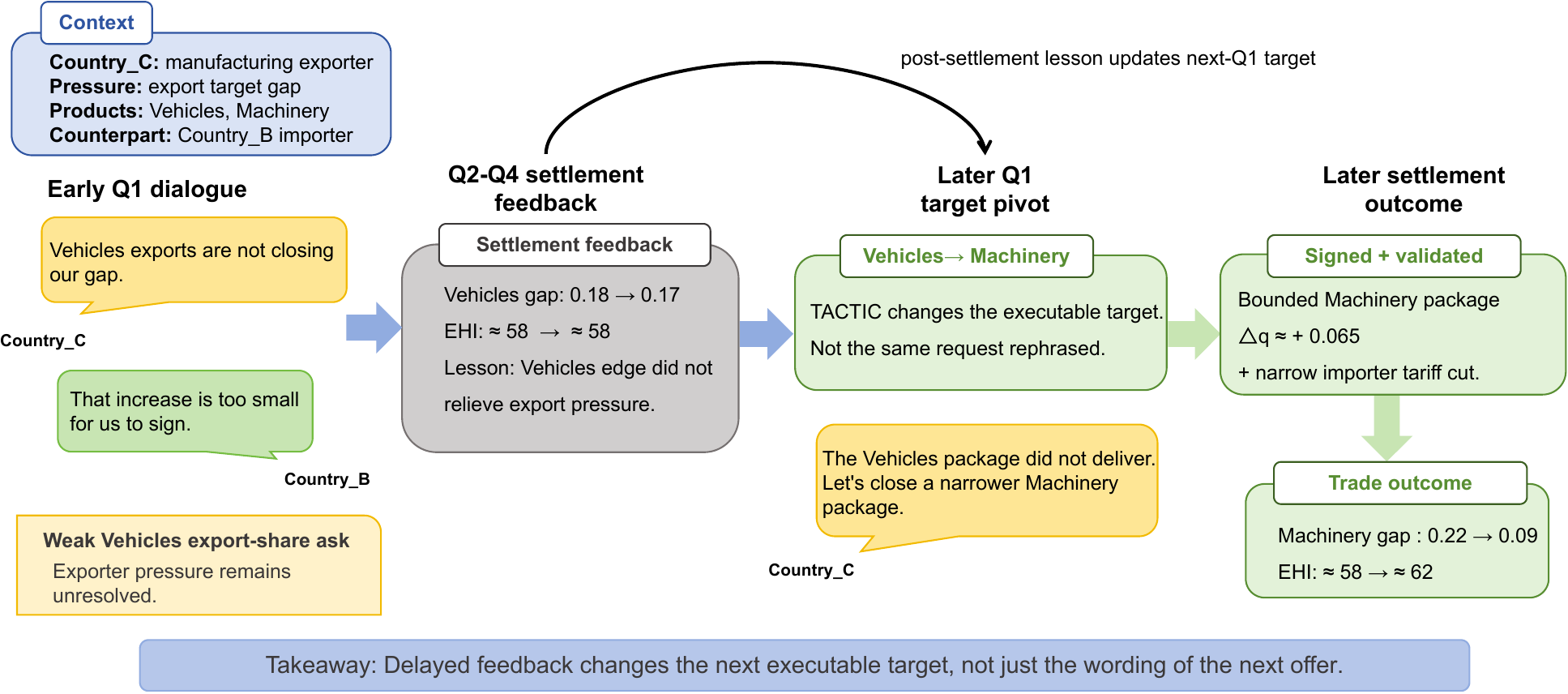}
\caption{Successful adaptation under export-target pressure. Early Vehicles bargaining yields little relief; after feedback, TACTIC pivots to a bounded Machinery package, reducing the Machinery target-gap ratio ($0.22\rightarrow0.09$) and raising EHI ($\approx 58\rightarrow 62$).}
\Description{A narrative timeline contrasting an earlier and later yearly episode for Country C negotiating with Country B under export-target pressure. The figure emphasizes the motivation for adaptation, the key dialogue showing that the Vehicles package is too weak, the settlement feedback that confirms little export relief, the next-Q1 pivot to a bounded Machinery package, and the improved post-settlement outcome.}
\label{fig:app_case_b1}
\end{figure*}

\subsection{Successful Adaptation: Rewriting the Next Executable Target}
\label{app:case_adaptation}

\paragraph{Scene.}
TACTIC controls \texttt{Country\_C}, a manufacturing hub whose scenario-defined role exports finished goods such as \textbf{Vehicles} and \textbf{Machinery}. Under the 20-episode sequential schedule from RQ2, with CoT-prompted \texttt{PromptLLM} opponents fixed across episodes, \texttt{Country\_C} faces persistent \emph{export}-target pressure on the bilateral edge with \texttt{Country\_B} (a resource/agriculture counterpart that imports those manufactures). The contrasted slices share similar opening pressure: Vehicles export target-gap ratio near $0.18$--$0.19$, limited room for an oversized share ask, and an importer unwilling to lock a large move without reciprocal value.

\paragraph{Dual-track timeline.}
Figure~\ref{fig:app_case_b1} isolates the decision-relevant moments. In the early episode, Q1 dialogue shows that the original \textbf{Vehicles} share request is too weak for the importer to accept, and Q2--Q4 settlement confirms little export relief. TACTIC uses this feedback in the later episode to change the target at the next Q1 window. It proposes a narrower, executable \textbf{Machinery} package consisting of a bounded share adjustment and a limited importer-side tariff cut by \texttt{Country\_B}.

\paragraph{Dialogue hinge.}
The hinge turn is \texttt{Country\_B}'s rejection of the Vehicles share request, followed by \texttt{Country\_C}'s shift to Machinery terms at the later Q1 window (Figure~\ref{fig:app_case_b1}).

\paragraph{Economic outcome.}
On the early-episode path, Vehicles remains unsigned or only symbolically adjusted; after Q1 application and Q2--Q4 settlement the Vehicles export target-gap ratio moves only $0.18\rightarrow0.17$, and country EHI ($\times 100$) stays near $58$. On the later-episode path, the Q1 proposal record does not reopen the stalled Vehicles ask: a Machinery export-share adjustment of about $\Delta q \approx +0.065$ on the C$\rightarrow$B edge, paired with a narrow importer-side tariff reduction by \texttt{Country\_B}, is signed, validated, and applied in Q1, then settled through Q2--Q4. The Machinery export target-gap ratio falls $0.22\rightarrow0.09$, and country EHI rises by about $+4$ points ($58\rightarrow 62$) over that year slice.

\paragraph{Why this path succeeds.}
Formal renegotiation cannot reopen in Q2--Q4, so adaptation occurs across negotiation windows. Post-settlement feedback from the early year shows that the Vehicles edge produced no material export-target relief. The next Q1 intent plan consequently changes both the \emph{product edge} and the \emph{structure} of the executable offer. After validation, the signed share and tariff terms alter environment state and affect quarterly settlement.

\paragraph{Insight.}
The case links the RQ2 trajectory to a specific behavioral change: settlement feedback alters the \emph{next} executable target. CoT-prompted memory or static variants can repeat a previously plausible package, whereas full TACTIC carries the observed delivery failure into the next negotiation window.

\begin{figure*}[t]
\centering
\includegraphics[width=0.95\linewidth]{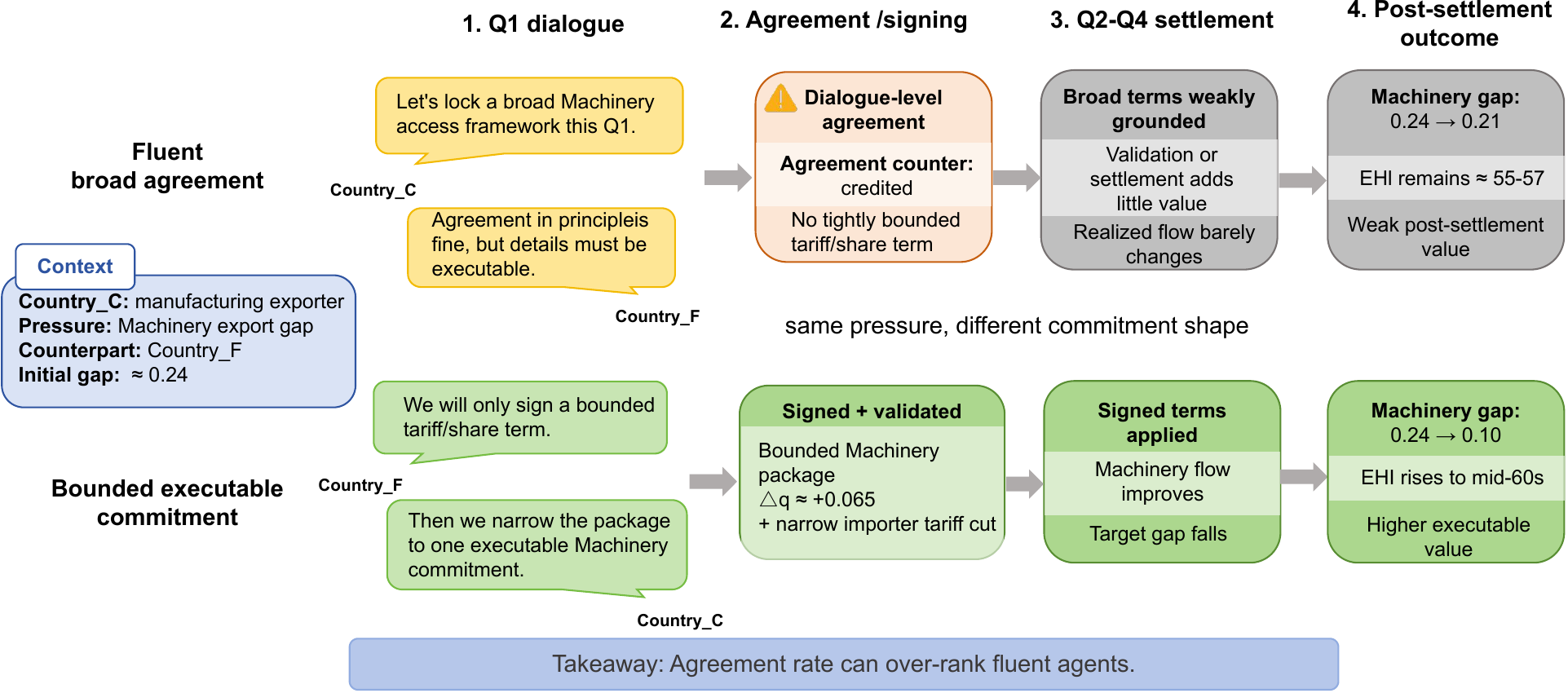}
\caption{Failure analysis: agreement versus post-settlement value. Broad Machinery access language yields little validated value ($0.24\rightarrow0.21$ gap), whereas a bounded export-share/tariff package executes and reduces the gap ($0.24\rightarrow0.10$).}
\Description{A two-path comparison of Machinery negotiation under the same export pressure. The upper path reaches broad verbal agreement but produces little post-settlement improvement, leaving the export target gap only slightly reduced and EHI nearly flat. The lower path signs a narrower executable package, after which settlement substantially lowers the target gap and raises country-level EHI.}
\label{fig:app_case_b2}
\end{figure*}

\subsection{Failure Case: Agreement Does Not Become Outcome}
\label{app:case_agreement_outcome}

\paragraph{Scene.}
The same manufacturing-export pressure appears in a cross-play slice where a fluent bargainer controls \texttt{Country\_C} against \texttt{Country\_F}. Both paths open a Q1 negotiation window with a large Machinery \emph{export} target-gap ratio ($\approx 0.24$) and cooperative public dialogue. They differ in whether the conversational consensus survives grounding and improves post-settlement health over Q1--Q4.

\paragraph{Dual-track timeline.}
Figure~\ref{fig:app_case_b2} contrasts a high-agreement path with an execution-viable path within the same calendar year. The upper track reaches a broad verbal deal in Q1; the lower track rejects or rewrites terms until the package is a validated export-share/tariff commitment narrow enough to move the gap through settlement.

\paragraph{Dialogue hinge.}
\texttt{Country\_F} conditions signing on bounded tariff/share terms after \texttt{Country\_C} proposes broad Machinery access. The lower-track response then specifies a share adjustment and tariff concession (Figure~\ref{fig:app_case_b2}).

\paragraph{Economic outcome.}
The fluent path records a dialogue-level success and is therefore counted by agreement rate. After grounding and Q1--Q4 settlement, however, the broad package is either discarded by validation or too weak to bind realized export flows: the Machinery export target-gap ratio moves only from $0.24$ to $0.21$, export-target completion stays low, and country EHI ($\times 100$) remains near $55$--$57$. On the execution-aware path, the signed terms reduce the gap ratio from $0.24$ to $0.10$ and raise country EHI to the mid-$60$s over the same year. This pattern provides trace-level context for Table~\ref{tab:baseline_comparison}: even with CoT prompting, \texttt{PromptLLM} reaches $75\%$ agreement but only 57.1 EHI.

\paragraph{Why this path fails.}
The failure occurs after substantive negotiation. Conversational consensus can fail to produce an environment-level consequence through a \emph{superficial agreement trap}, in which language does not become a validated tariff/share term, or an \emph{execution-binding failure}, in which valid terms deliver insufficient realized trade, for example under capacity rationing. Evaluation at signing would score the fluent track without observing either failure.

\begin{figure*}[t]
\centering
\includegraphics[width=0.95\linewidth]{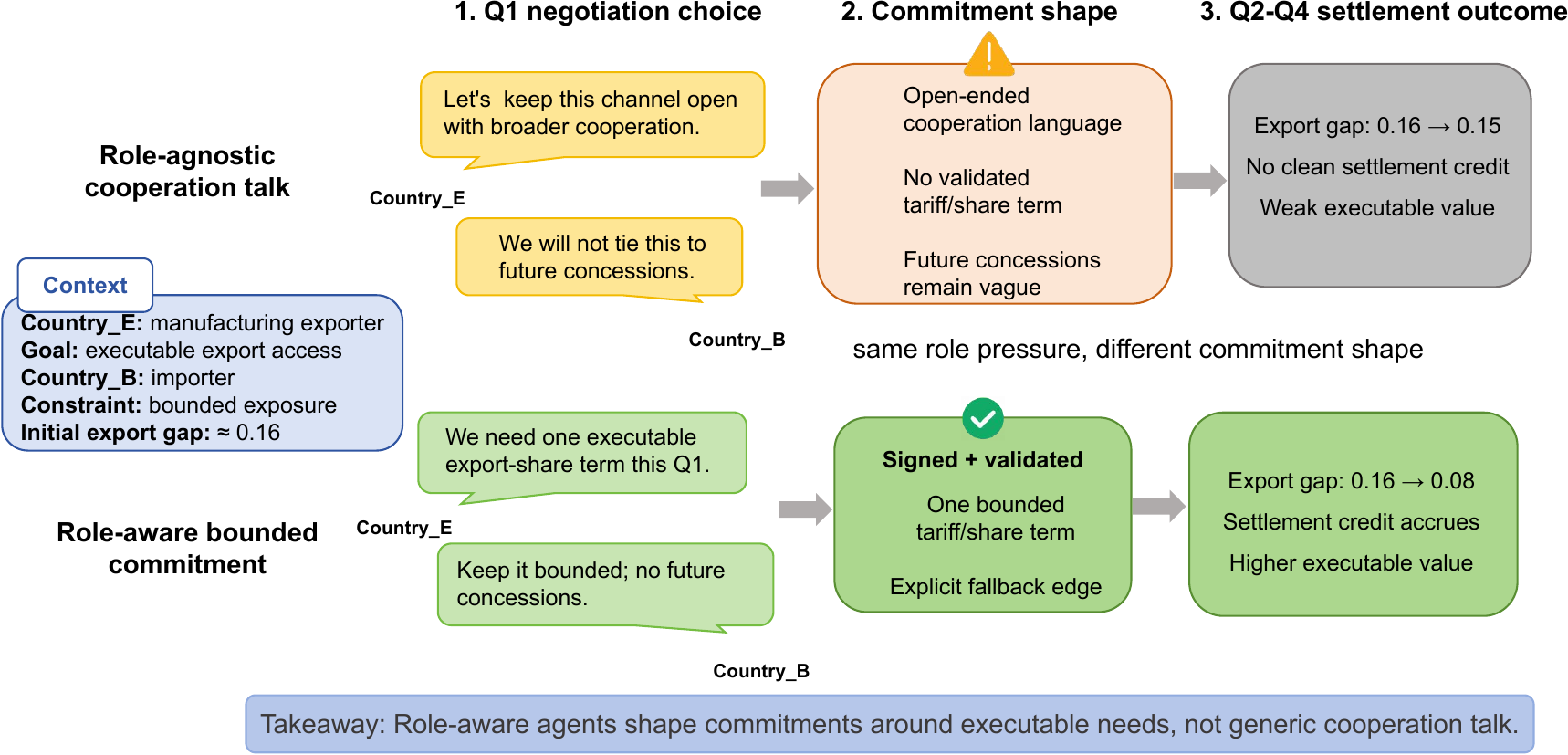}
\caption{Role-conditioned strategy under asymmetric trade pressure. Broad cooperation barely moves the active-edge gap ($0.16\rightarrow0.15$), while a bounded tariff/share commitment with fallback reduces it ($0.16\rightarrow0.08$).}
\Description{A two-path comparison showing role-conditioned negotiation under asymmetric trade pressure. The upper path relies on broad cooperative language and loosely framed future concessions, leading to little movement in the export target gap. The lower path signs one bounded executable commitment with an explicit fallback, after which settlement cuts the gap roughly in half without expanding into a broader unenforceable agreement.}
\label{fig:app_case_b3}
\end{figure*}

\paragraph{Insight.}
The case locates the performance difference after dialogue-level agreement. An agent may produce agreeable text without producing a commitment that improves delayed macroeconomic outcomes. Post-execution EHI captures this difference, which agreement rate alone omits.

\subsection{Role-Conditioned Strategy: Exporter Access vs. Importer Exposure}
\label{app:case_bounded}

\paragraph{Scene.}
\texttt{Country\_E}, an emerging manufacturing exporter controlled by TACTIC, faces \texttt{Country\_B}, an importer whose prior counters are volatile and whose exposure to future concessions must be bounded. The role pressure is asymmetric: \texttt{Country\_E} needs executable export access on the active edge, whereas \texttt{Country\_B} will not lock broad future obligations. The case examines whether the resulting Q1 commitment reflects both countries' executable constraints.

\paragraph{Dual-track timeline.}
Figure~\ref{fig:app_case_b3} contrasts a role-agnostic and a role-aware path under the same export pressure. The upper track uses broad cooperation language to keep the channel open, but it does not reconcile the exporter's need for access with the importer's reluctance to accept unbounded future obligations. The lower track narrows the exchange into one limited tariff/share term in Q1 and identifies a fallback product or partner edge if the counterpart refuses, to be pursued at a later negotiation window.

\paragraph{Dialogue hinge.}
\texttt{Country\_E} requests one executable export-share term, and \texttt{Country\_B} rejects unbounded future concessions. The response narrows the agreement to a limited tariff/share commitment with a fallback product or partner edge (Figure~\ref{fig:app_case_b3}).

\paragraph{Economic outcome.}
On the broad-cooperation track, the parties exchange cooperative assurances, but the proposal record either lacks a validated bilateral term or pairs soft language with an oversized ask that the counterpart will not lock; the active-edge export target-gap ratio changes only from $0.16$ to $0.15$ over Q2--Q4. On the bounded track, a single signable term is recorded, validated, and applied in Q1. Settlement reduces the gap ratio to $0.08$, and later proposal steps within that window do not expand the agreement beyond the counterpart's stated limit.

\paragraph{Why the strategies differ.}
Role pressure changes the \emph{structure} of the commitment. The exporter requires a concrete access term that can reduce the target gap, while the importer requires bounded exposure. Broad cooperation language does not resolve these constraints; bounded tariff/share terms do. A fallback is operationally relevant when refusal redirects the agenda to another edge at a future Q1 window, because settlement-only quarters do not permit continuous renegotiation.

\paragraph{Insight.}
This case tests role-conditioned commitment discipline: selecting a smaller enforceable exchange that fits both countries' constraints instead of a broader conversational coalition. TradeWar measures the resulting difference because unvalidated language leaves state unchanged, whereas a signed policy updates the environment.

Across the three cases, signed terms alter economic state, dialogue-level agreements can fail during validation or settlement, and role constraints determine the value of a bounded commitment. These trace patterns explain the aggregate difference between agreement rate and EHI. They also show how inference-time adaptation operates in TradeWar: delayed feedback revises subsequent executable targets and package structure.

\end{document}